\documentclass{article} 
\usepackage{collas2023_conference,times}
\usepackage{easyReview}

\usepackage{amsmath,amsfonts,bm}

\def\eqref#1{equation~\ref{#1}}

\def\1{\bm{1}}

\DeclareMathAlphabet{\mathsfit}{\encodingdefault}{\sfdefault}{m}{sl}
\SetMathAlphabet{\mathsfit}{bold}{\encodingdefault}{\sfdefault}{bx}{n}

\usepackage{hyperref}
\hypersetup{
    colorlinks=true,
    linkcolor=red,
    filecolor=magenta,
    urlcolor=blue,
    citecolor=purple,
    pdftitle={Overleaf Example},
    pdfpagemode=FullScreen,
    }

\usepackage{bbm}
\usepackage{amsmath}
\usepackage{soul}
\sethlcolor{red}
\usepackage{url}
\usepackage[utf8]{inputenc} 
\usepackage[T1]{fontenc}    
\usepackage{hyperref}       
\usepackage{url}            
\usepackage{booktabs}       
\usepackage{amsfonts}       
\usepackage{nicefrac}       
\usepackage{microtype}      
\usepackage{xcolor}         
\usepackage{graphicx}
\usepackage{subcaption}
\usepackage{amsmath}
\usepackage{amssymb}
\usepackage{mathtools}
\usepackage{amsthm}

\usepackage{amssymb}
\usepackage{dirtytalk}
\usepackage{float}

\usepackage{multicol}
\usepackage{authblk}
\usepackage{tikz}
\usepackage{amsmath}
\usepackage{pythonhighlight}

\usepackage{amsmath}
\usepackage{amssymb}
\usepackage{mathtools}
\usepackage{amsthm}
\usepackage{algorithm}
\usepackage{algorithmic}
\usepackage{wrapfig}
\usepackage{enumitem}
\setlist[itemize]{leftmargin=1.5mm}

\usepackage[capitalize,noabbrev]{cleveref}

\theoremstyle{plain}

\theoremstyle{definition}

\theoremstyle{remark}

\newcommand{\modif}[1]{\textcolor{black}{#1}}

\newcommand{\Scole}{\textit{SCoLe}}

\title{Challenging Common Assumptions about Catastrophic Forgetting}

\author{Timothee Lesort$^{12}$ Oleksiy Ostapenko$^{12}$ Diganta Misra$^{12}$  Md Rifat Arefin$^{12}$  Pau Rodr\'iguez$^3$ Laurent Charlin$^{145}$ Irina Rish$^{125}$
    \\ 
    \small$^1$Mila - Quebec AI Institute, $^2$Université de Montréal, $^3$Apple,  $^4$HEC Montréal, \\ $^5$Canada CIFAR AI Chair  \\}

\preprintcopy 

\begin{document}

\maketitle

\begin{abstract}

Building learning agents that can progressively learn and accumulate knowledge is the core goal of the continual learning (CL) research field. 
Unfortunately, training a model on new data usually compromises the performance on past data.
In the CL literature, this effect is referred to as catastrophic forgetting (CF).
CF has been largely studied, and a plethora of methods have been proposed to address it on short sequences of non-overlapping tasks. In such setups, CF always leads to a quick and significant drop in performance in past tasks.
Nevertheless, despite CF, recent work showed that SGD training on linear models accumulates knowledge in a CL regression setup. This phenomenon becomes especially visible when tasks reoccur.
We might then wonder if DNNs trained with SGD or any standard gradient-based optimization accumulate knowledge in such a way.
Such phenomena would have interesting consequences for applying DNNs to real continual scenarios. Indeed, standard gradient-based optimization methods are significantly less computationally expensive than existing CL algorithms.
In this paper, we study \modif{the progressive} knowledge accumulation (KA) in DNNs trained with gradient-based algorithms \modif{in long sequences of tasks} with data re-occurrence. 
We propose a new framework, \Scole{} (Scaling Continual Learning), to investigate KA and discover that catastrophic forgetting has a limited effect on DNNs trained with SGD. When trained on long sequences with data sparsely re-occurring, the overall accuracy improves, which might be counter-intuitive given the CF phenomenon. 
We empirically investigate KA in DNNs under various data occurrence frequencies and propose simple and scalable strategies to increase knowledge accumulation in DNNs.

\end{abstract}



\section{Introduction}
\label{sec:Introduction}
\begin{wrapfigure}[18]{R}{0.45\linewidth}
    \centering  
    \includegraphics[width=0.9\linewidth]{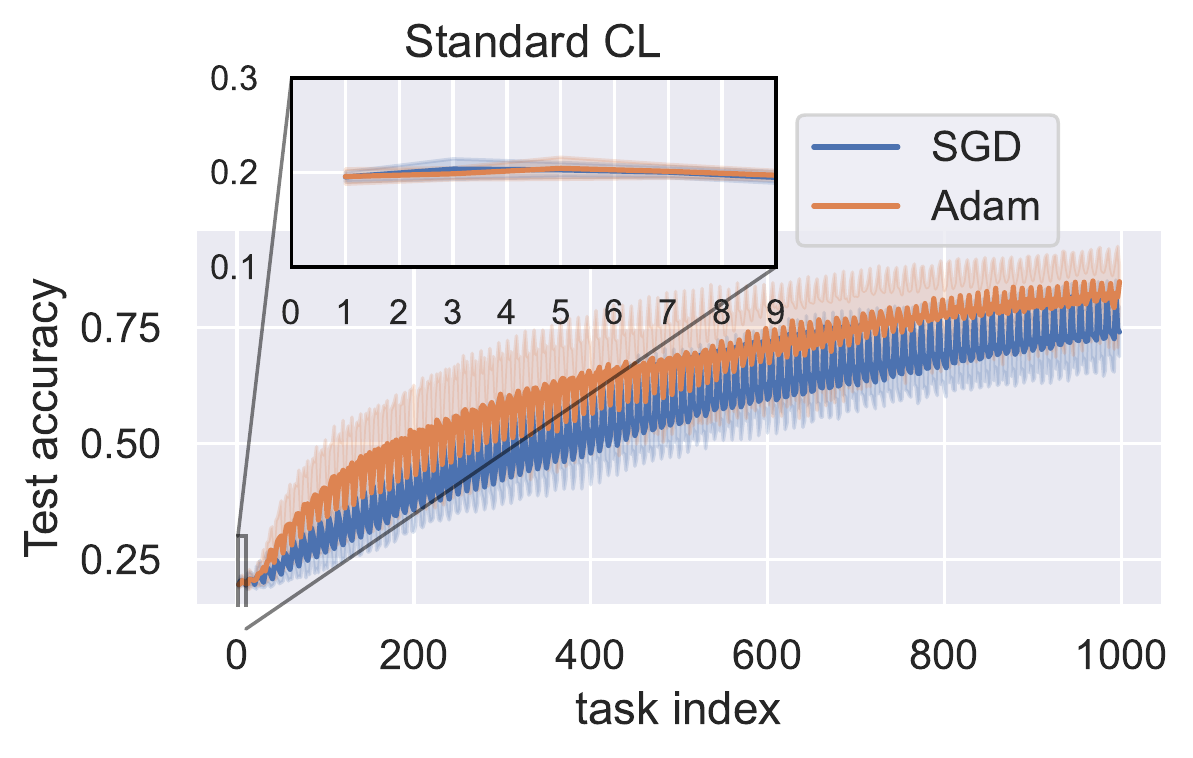}
    \caption{Knowledge accumulation is not observable in standard CL benchmarks (inset top). However, when repeating the sequence of tasks (bottom), it becomes apparent (MNIST, 2 classes per task, averaged over 3 lr and 5 seeds, each task until convergence)}
    \label{fig:fig_1}
\end{wrapfigure} 

Continual learning (CL) aims to design algorithms that learn from non-stationary sequences of tasks and accumulate knowledge. Classically, the main challenge of CL is catastrophic forgetting (CF) --- fast performance degradation on previous tasks when learning from new data. 
CF is evaluated based on the 0-shot accuracy of past tasks in scenarios with short sequences of disjoint tasks~\citep{Lesort2019Continual,deLange2019continual,BELOUADAH202138,hadsell2020embracing}. 
This evaluation protocol has led our community to the conclusion that new knowledge systematically replaces the previous one, hence that catastrophic forgetting erases past knowledge.

We challenge this idea and step back from classical CL and CF literature to investigate to which extent fine-tuning with SGD under sequential distribution shifts prevents the model from accumulating knowledge. 

Our research is driven by a simple observation highlighted in Fig.~\ref{fig:fig_1}. It shows that a DNN trained with SGD and Adam  accumulates knowledge after repeated exposure to a fixed sequence of tasks, even in a single-head evaluation mode (i.e., without a separate head per task), which contradicts the dominant belief in the CL literature that CF should prevent such accumulation. To investigate this phenomenon further, we demonstrate in \cref{sec:knowledge_retention} that a DNN can retain knowledge about a one task even after several learning unrelated tasks (which we validate using meta-testing accuracy).

Motivated by these observations, we investigate the effect of knowledge accumulation in long sequences of tasks. We propose \Scole{} (Scaling Continual Learning), a framework that allows the generation of long sequences of potentially overlapping tasks (\cref{fig:illustration}). 
\Scole{} relaxes the constraint that tasks cannot overlap but constrains the sampling to a subset of classes that changes over time  (locally stationary). Samples are non-IID, and learning a task can still lead to CF of previous tasks. 
Thus, Scole is designed to reveal whether proper knowledge accumulation occurs in models or if forgetting prevents it.
Our findings reveal that when scaled to hundreds or thousands of tasks, deep neural networks (DNNs) trained with stochastic gradient descent (SGD) display a consistent accumulation of knowledge, providing evidence that the impact of catastrophic forgetting is limited. 
We study the impact of class and task occurrence frequencies on model performance to understand better the knowledge accumulation capability of deep neural networks. We also explore different options to boost knowledge accumulation and benefit from it.

Our contributions are as follows: 

\begin{itemize} 
    \item We first show that forgetting does not prevent the transfer to downstream tasks, even several tasks later. We call this effect \say{knowledge retention}, and we show it leads to knowledge accumulation when tasks are reoccurring.
    \item We propose an experimentation framework \say{\Scole{}} (\textbf{S}caling \textbf{Co}ntinual \textbf{Le}arning) to study and assess knowledge accumulation through a long sequence of (reoccurring) tasks on a variety of datasets (MNIST, Fashion-MNIST, KMNIST, CIFAR10, CIFAR100, Tiny-ImageNet200).
    \item Our study reveals that deep neural networks (DNNs) trained with SGD on a non-stationary distribution are capable of accumulating knowledge, challenging the usual understanding of the impact of catastrophic forgetting.
    \item We investigate different strategies to improve knowledge accumulation effects, such as gradient masking, frequency replay, optimizing the learning-forgetting trade-off or making models wider.
\end{itemize}

\section{The Long-Term Effect of Forward Transfer} 
\label{sec:knowledge_retention}

\modif{In the continual learning literature, transfer is a major axis of investigation alongside CF \citep{lopezpaz2017gem,riemer2018learning}. The transfer can either be forward when already learned tasks help to solve the following ones, or backward when learning new tasks helps with solving past tasks \citep{Lin2022BeyondNC}. Forward transfer is usually evaluated through the comparison of performance on individual tasks trained IID to their performance obtained in CL streams~\citep{lopezpaz2017gem,veniat2021efficient}. Recently, one of the major types of transfer investigated in the literature is the transfer from pre-trained models to a sequence of downstream tasks \citep{Ostapenko2022Foundational,Hayes202REMIND,ke2023continual}. The transfer is usually evaluated through 0-shot performance, but can also be evaluated by probing ~\citep{davari2022probing,caccia2019online, fini2020online}. 
Even if the literature investigating large pre-trained models gives clear insight into forward transfer having effects on all the downstream tasks of a continual learning scenario, forward transfer is usually evaluated by looking at how learning one task influences the following one without investigating explicitly effects on a longer term.}
\begin{wrapfigure}[14]{l}{0.25\textwidth}
  \centering                                
  \includegraphics[width=1\linewidth]{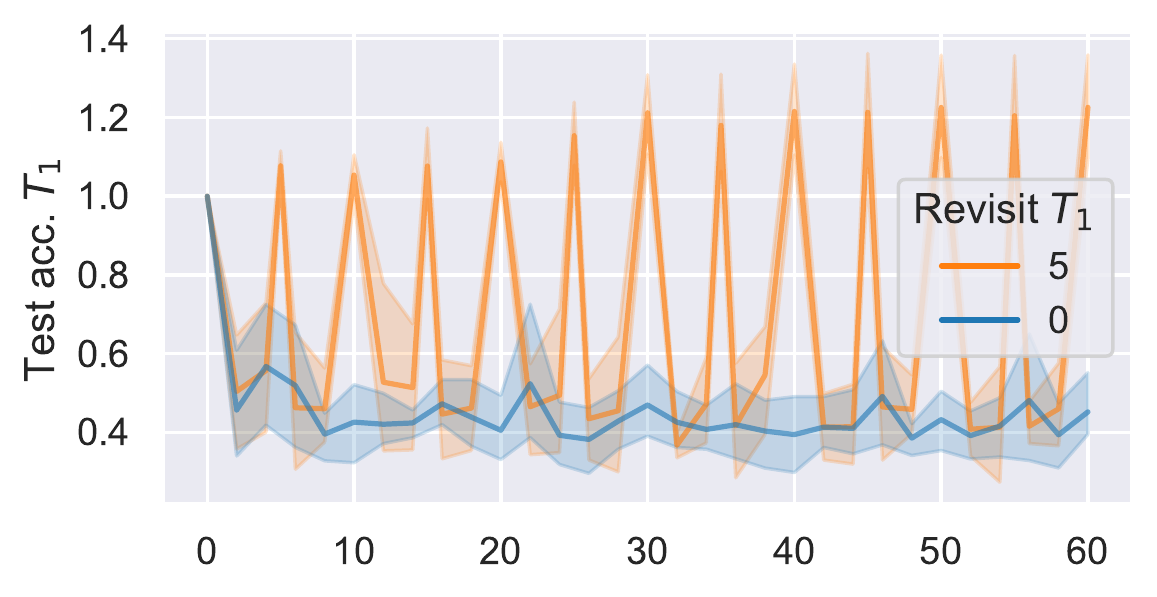}      
  \caption{0-shot test acc. on task 1 while learning distractor tasks and revisiting task 1 at different frequencies (i.e. every 5 tasks or never). This plot suggests that the model completely forgets between two occurrences of task 1. 
  }
  \label{fig:knowledge_transfer}
\end{wrapfigure}

\modif{In this section, we investigate if forward transfer can have a long-term impact beyond the direct next task. From a different perspective, we evaluate if the model can retain knowledge (knowledge retention -- KR) over extended periods, and whether it can accumulate it (knowledge accumulation -- KA), in non-stationary training regimes.}

\modif{We design a scenario with one task of interest (task 1) that reoccurs at a regular fixed frequency with intermediate distractor tasks (we run for 5 different seeds, each resulting in task 1 composed of different randomly selected classes). The goal of this setting is to track if the distractor tasks lead to complete forgetting or not, and if not, how catastrophic the forgetting is.
The scenario is designed from CIFAR100, the task of interest is a classification task with 5 randomly selected classes. Distractors tasks are built by sampling 5 random other classes and applying pixel-level permutation to the original images (akin to permuted MNIST~\citep{goodfellow2013empirical}).}

\modif{We train a model with ResNet architecture, consequently since CNNs usually rely on features related to neighbouring pixels to learn, the pixel permutation of the distractor tasks should be especially perturbing and should lead to forgetting. The model is trained until convergence on each task.}

\modif{We measure both the test and meta-test performance of the incrementally trained  model. Test performance denotes the usual test accuracy directly measured on the trained model (i.e. 0-shot performance) while meta-test performance refers to the accuracy attained by the model if finetuned for a single epoch on task 1 (the task of interest).
The meta-test measures how the model is potentially able to transfer knowledge from the past task of interest to a new instance of the same task.}

\begin{wrapfigure}[15]{r}{0.25\textwidth}
  \centering                                                 
  \includegraphics[width=1\linewidth]{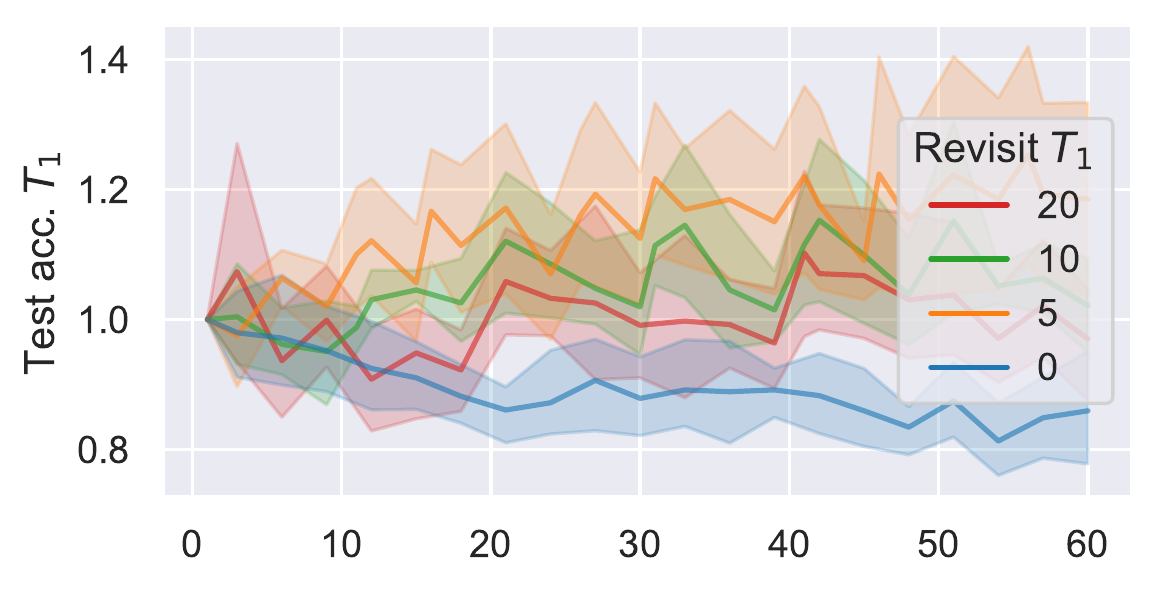}      
  \caption{Normalized meta-test accuracy on task 1 while learning distracting tasks and revisiting task 1 at different frequencies. 
  The meta-test accuracy shows that beyond 0-shot forgetting, the model performs knowledge retention and accumulation.
  }
  \label{fig:knowledge_retention}
\end{wrapfigure}

\modif{\cref{fig:knowledge_transfer}, shows a clear catastrophic forgetting behaviour in 0-shot performance. In this setting, distractor tasks make the model completely forget in terms of 0-shot performance, even if task 1 is revisited every 5 tasks. We will see now if the forgetting is that dazzling for meta-testing or if, in fact, the model retains knowledge from the previous task and can still transfer it.}
\modif{If the transfer had limited long-term impact, we would expect the meta-test performance to decrease very fast, while if the transfer had long-term impact the meta-test performance should decrease slowly. We can note that long-term transfer is equivalent to knowledge retention. Then, if in addition to knowledge retention, the model can perform knowledge accumulation, the meta-test performance should increase progressively when the task is revisited.}

\modif{\cref{fig:knowledge_retention} shows that meta-test accuracy actually slowly decreases over time when trained on distractor tasks, and has an upward tendency if task 1 is revisited regularly, with a faster increase for more frequent revisits. This shows that the model can indeed retain past knowledge for prolonged periods and accumulate it, even if the 0-shot test accuracy shows catastrophic forgetting and abrupt drops as in ~\cref{fig:knowledge_transfer}.
Note: In~\cref{fig:knowledge_retention} and~\cref{fig:knowledge_transfer}, we experiment with various sets of classes for task 1, which lead to variability in results depending on the difficulty of this task, to reduce the variance, we normalize accuracies by the task 1 accuracy at its first occurrence.}

\modif{\textbf{Summary.} In this section, we showed that despite 0-shot catastrophic forgetting, the model retains knowledge over extended periods of time, even after training on several distractor tasks. Moreover, when one task is revisited periodically, the model progressively accumulates knowledge. In this setup, the effect of knowledge accumulation is visible when probing the model (meta-test) but not in 0-shot performance.
However, in many practical settings, one is interested in a 0-shot performance of the overall problem and not in the meta-testing performance. Moreover, to increase task variability and only target scalable approaches, we are interested in setting where data do not reoccur in exactly the same context (the same task) and where the number of training stages (tasks) is high (>500 tasks).
KA could also have consequences in 0-shot performance when scaling the number of tasks and repetitions.
Motivated by this, in the later sections, we propose the \Scole{} framework, which implements those requirements and enables us to study if knowledge accumulation in DNNs leads to a progressive improvement of 0-shot accuracy in continual learning scenarios.}

\section{\Scole{}:A Framework for CL with long Task Sequences}
\label{sec:framework}

 \begin{figure}[ht]
    \centering             
    \includegraphics[width=\linewidth]{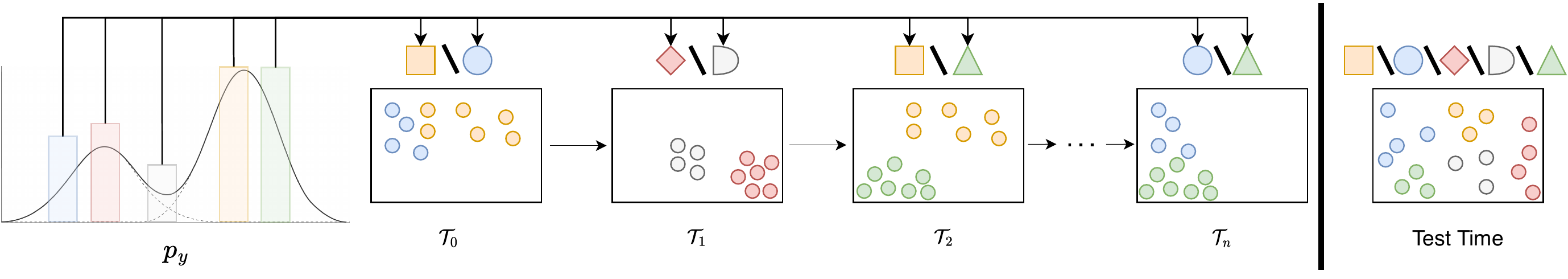}
    \caption{Illustration of \Scole{} (Scaling Continual Learning) scenario. With 5 classes in total (one per colour) and 2 classes per task. The data is selected randomly based on their label to build tasks dynamically into a potentially infinite sequence. The model is trained on each task and evaluated on the test set containing all possible classes.}
    \label{fig:illustration}
\end{figure} 

To study knowledge accumulation in long sequences of tasks, we propose the \Scole{} (Scaling Continual Learning) framework. 
This framework enables the creation of scenarios with an arbitrarily long sequence of tasks. As in classical CL, each task's training set in \Scole{} contains a subset of a dataset. The evaluation is conducted by measuring the overall performance of this dataset. In this setting, a learning system must accumulate knowledge by sequentially experiencing isolated parts of the data. The key difference from classical CL scenarios is that data in \Scole{} can reoccur in different contexts.

Modern deep learning systems demonstrate the capability to accumulate knowledge when trained on stationary data. Nonetheless, in situations where data from a shifted distribution is introduced incrementally and does not re-occur, CF hinders the accumulation of knowledge.
\Scole{} proposes to fill the gap between these two settings and to study how knowledge accumulation evolves in non-stationary regimes as re-occurrence becomes more sparse in time.
One of \Scole{} goals is to determine the occurrence frequency regimes in which algorithms can effectively learn. In this paper, the algorithms of interest are how gradient-based algorithms, such as SGD.
Intuitively, between two occurrences of a task, DNNs are trained on other tasks and may forget. 
We measure the test accuracy on the entire test set to track whether DNNs are learning more than forgetting. A model learning consistently more than it forgets will progressively converge to a solution for all tasks.
Nevertheless, the trade-off between learning and forgetting depends on many factors, notably the occurrence frequency of the data.
Experimenting with \Scole{} makes it possible to vary the frequency occurrence to understand better the learning dynamics of a given approach. 

\modif{As noted in \cite{cossu2021classincremental} the reocurrence of data happens naturally in many settings, and continual learning algorithms can benefit from it.}
 One example of a real scenario where data reoccurs is robotics. In a robotics environment, such as a building or a factory, the tasks and concepts may vary through time but regularly reoccur. It is then important to know under which task re-occurrence frequencies algorithms can be successful or not to be able to improve them. Other examples could also be to streaming setups where events can repeat, e.g. newspaper feed, social feed, or stock market...
Intuitively, the training environment that fits \Scole{} are those where the finite world assumption  \citep{mundt2020wholistic} may hold and where training is realized over a long period of time.

\textbf{Framework.}   
We instantiate \Scole{} in a classification setting  (\cref{fig:illustration}). 
In \Scole{}, a task is a period of time where the data distribution is stationary. The task changes with the input distribution. The test set stays fixed. 
In more detail, each task consists of samples from a random subset of the $N$ total available classes ($N$ is dataset dependent). The agent is a DNN that trains on the current task but evaluates on the test set $D_{test}$ containing the data of all classes. The framework considers scenarios with varying numbers of tasks $T$ and classes per task $C$. 
 
Formally, the training set $D_{t}$ for a task $t$ consists of observations $(x,y)$, which are generated like: $y\sim U(S_t)$, $x\sim p(X|Y=y)$. Here, the set $S_t=\{c_i\}_{i=0}^{C-1}$ is sampled without replacement from $P_{\Scole{}}(Y)$ and contains classes of task $t$. $P_{\Scole{}}(Y)$ is a probability distribution defined over $N$ classes. 
We set $P_{\Scole{}}(Y)$ at the beginning to define the scenario and set the occurrence frequency of each class.

In the default \Scole{} scenario, $P_{\Scole{}}(Y)$ is the uniform distribution over all $N$ classes $U(0, N-1)$. 
We can also consider cases where $P_{\Scole{}}(Y)$ is non-uniform (\cref{sub:mixture}) or evolves over time (\cref{ap:forgetting}). 

 \modif{\Scole{} scenarios can also be built with additional constraints. We experimented with some ideas in the appendix, such as: penalizing reoccurrence of recent classes in \cref{ap:sub:prob_reduction}, restricting the number of possible tasks in \cref{ap:sub:pairs}, imposing fix sequence of tasks as in \cref{fig:fig_1} but with a random change in the classes in \cref{ap:sub:structure} or creating cyclic shifts in the distribution of classes in \cref{ap:sub:cyclic}. }

\textbf{Frequency of Occurrence. }
One of the key features of \Scole{} is the ability to study KA as a function of the frequency of occurrence ($\nu$) and determine the frequency range at which algorithms are efficient. Algorithms designed for IID settings are known to be effective learners when $\nu$ is very high ($\sim$ every batch). on the contrary, CL algorithms are designed for settings where $\nu$ is very low ($\sim$ for some batches and then never again)\citep{kirkpatrick2017overcoming}. Experimenting with \Scole{} will show performance in the intermediate frequencies of occurrence. 

We can compute the expected occurrence frequency $\nu^*_{class}$ of one class from $P_{\Scole{}}(Y)$, $N$ and $C$. 
 For example, for a uniform $P_{\Scole{}}(Y)$, the expected frequency of occurrence of a class can be calculated with the probability of sampling $C$ elements from a set of $N$ unique elements without replacement and can deduce the frequency of occurrence per task directly: 
 %
 $\nu^*_{class}= 1 - \Pi_{i=0}^{C-1} (1 - \frac{1}{N-i})$ per task ($task^-1$).
%
Then the expected number of tasks between two occurrences of a given class $\tau^*_{class}= \frac{1}{\nu_{class}}$. For example, with CIFAR10 the total number of classes is $10$, hence with $C=2$ and uniform $P_{\Scole{}}(Y)$, we have $\nu^*_{class}=1 - (1-\frac{1}{10})(1 - \frac{1}{9}) = 0.200~task^{-1}$ then $\tau_{class}=5$ tasks. 

%
 For the task occurrence frequency, the expected number of tasks between two occurrences of the same task depends on the total number of tasks ${N \choose C}$.
 With CIFAR100 and 2 classes per task, $\tau_{task}={100 \choose 2}=4950$. Thus, revisiting exactly the same task in this setup is very rare (every 4,095 tasks). Note for standard CL $\tau_{task}$ and $\tau_{class}$ are $\inf$. 
 In this paper, the occurrence of frequency is always the expected occurrence frequency (a priori set) and not the empirical frequency (measured by counting the number of occurrences).


\textbf{Summary.} \Scole{} is a continual learning framework for generating long sequences of tasks with various frequencies of tasks and classes. It is made to study the knowledge accumulation capability of learning algorithms.
\Scole{} scenarios are in between IID scenarios and usual CL scenarios, as the data may reoccur, but the distribution changes through time and forgetting can compromise performance. \Scole{} can be seen as a bridge between both setups since, depending on the frequency occurrence of classes, we can be closer to one or the other.

\section{Knowledge Accumulation}
\label{sec:preliminaries}

As documented in various classical works on CL, including \cite{French99}, CF typically leads to a decline in performance on previously learned tasks. Hence, we could expect that in any non-IID scenario, forgetting might have a strong effect on performance. Building upon the insights presented in Sec.~\ref{sec:knowledge_retention}, we investigate knowledge accumulation in a simple \Scole{} setup. 
In contrast to Sec.~\ref{sec:knowledge_retention}, where the task performance where measure independently for other tasks, we now measure the overall 0-shot performance (full test set) set with a single-head architecture.  

\subsection{Initial Experiments on \Scole{}}
\begin{figure}[tp]
    \begin{subfigure}[b]{\linewidth}
    \centering
    \includegraphics[width=\linewidth]{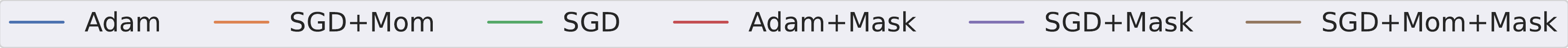}
    \end{subfigure}
    
    \begin{subfigure}[b]{0.24\linewidth}
    \centering
    \includegraphics[width=\linewidth]{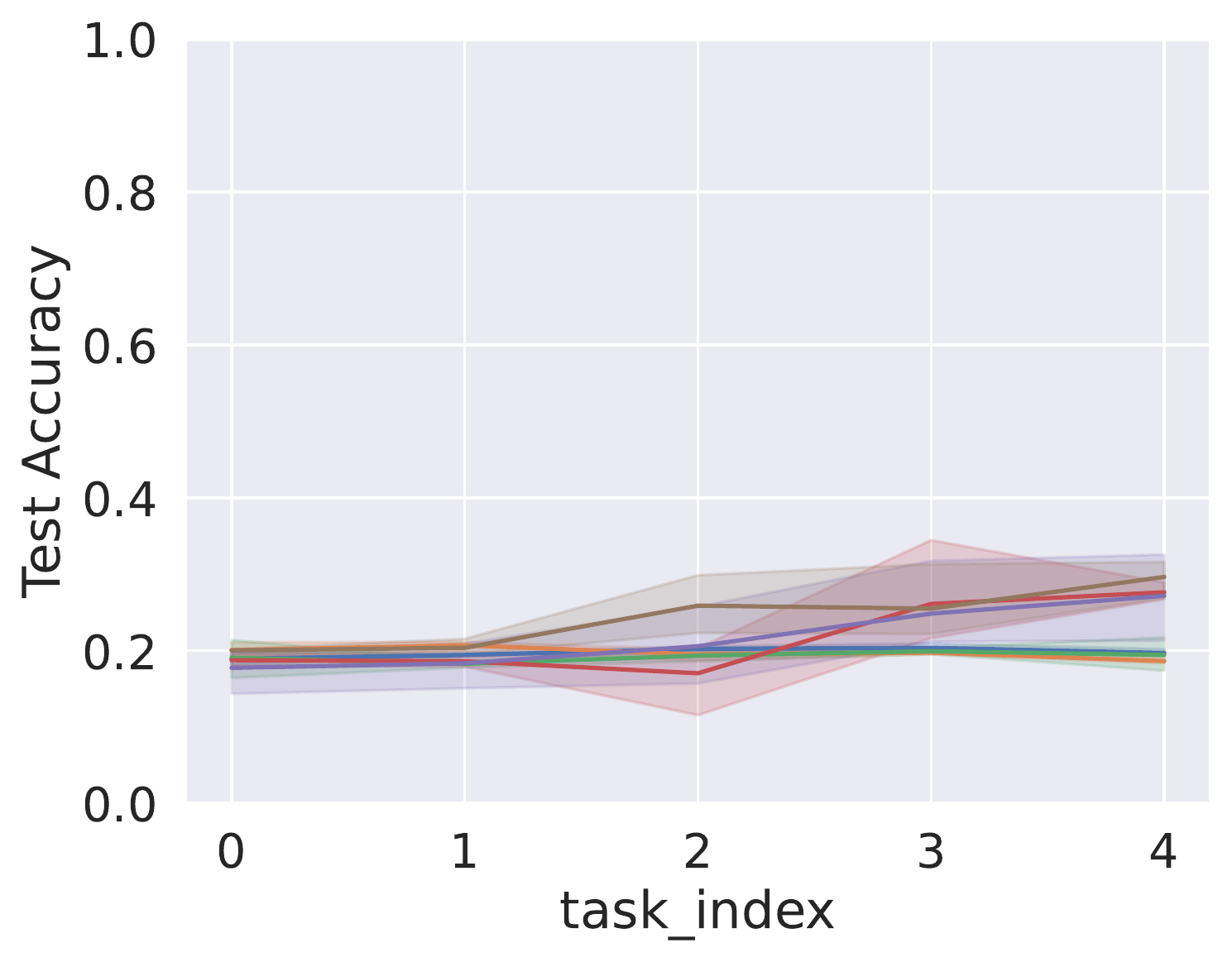}
    \caption{Short Sequence}
    \label{fig:first_imp}
    \end{subfigure}
    \begin{subfigure}[b]{0.24\linewidth}
    \centering
    \includegraphics[width=\linewidth]{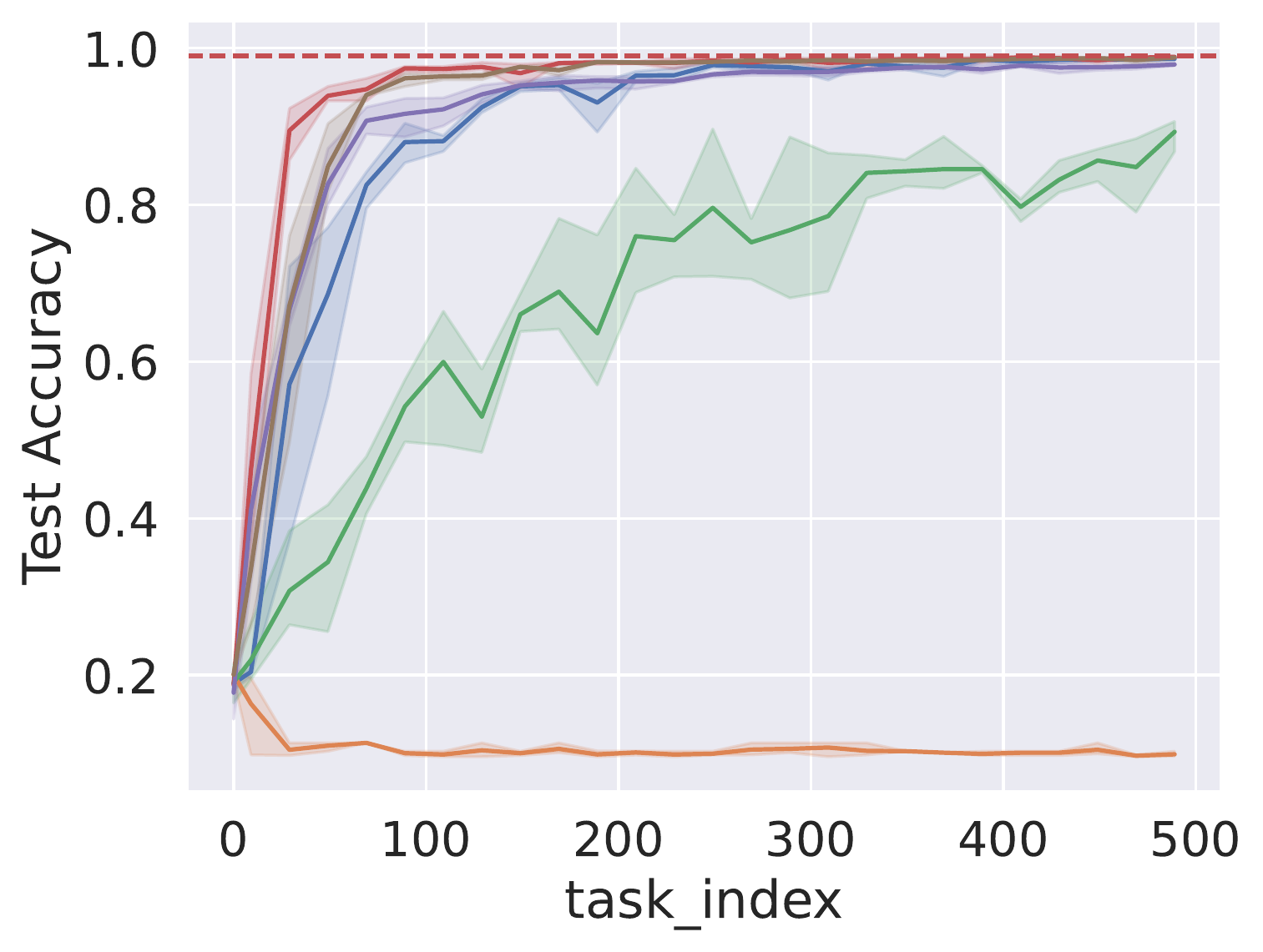}
    \caption{Scaling}     
    \label{fig:first_scaling}
    \end{subfigure}
    \begin{subfigure}[b]{0.24\linewidth}
    \centering
    \includegraphics[width=\linewidth]{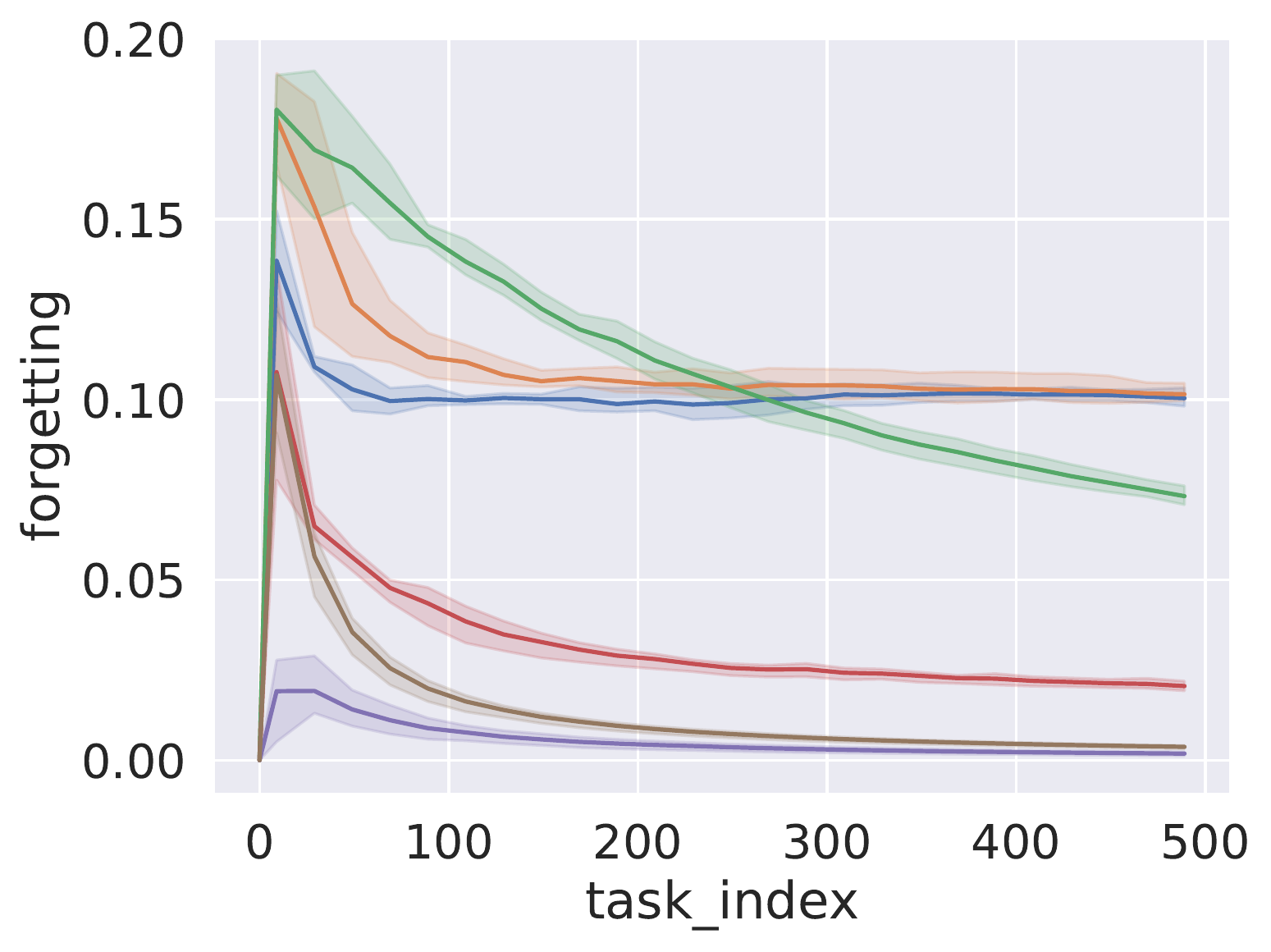}
    \caption{Forgetting}
    \label{fig:first_imp_fg}
    \end{subfigure}
    \begin{subfigure}[b]{0.24\linewidth}
    \centering
    \includegraphics[width=\linewidth]{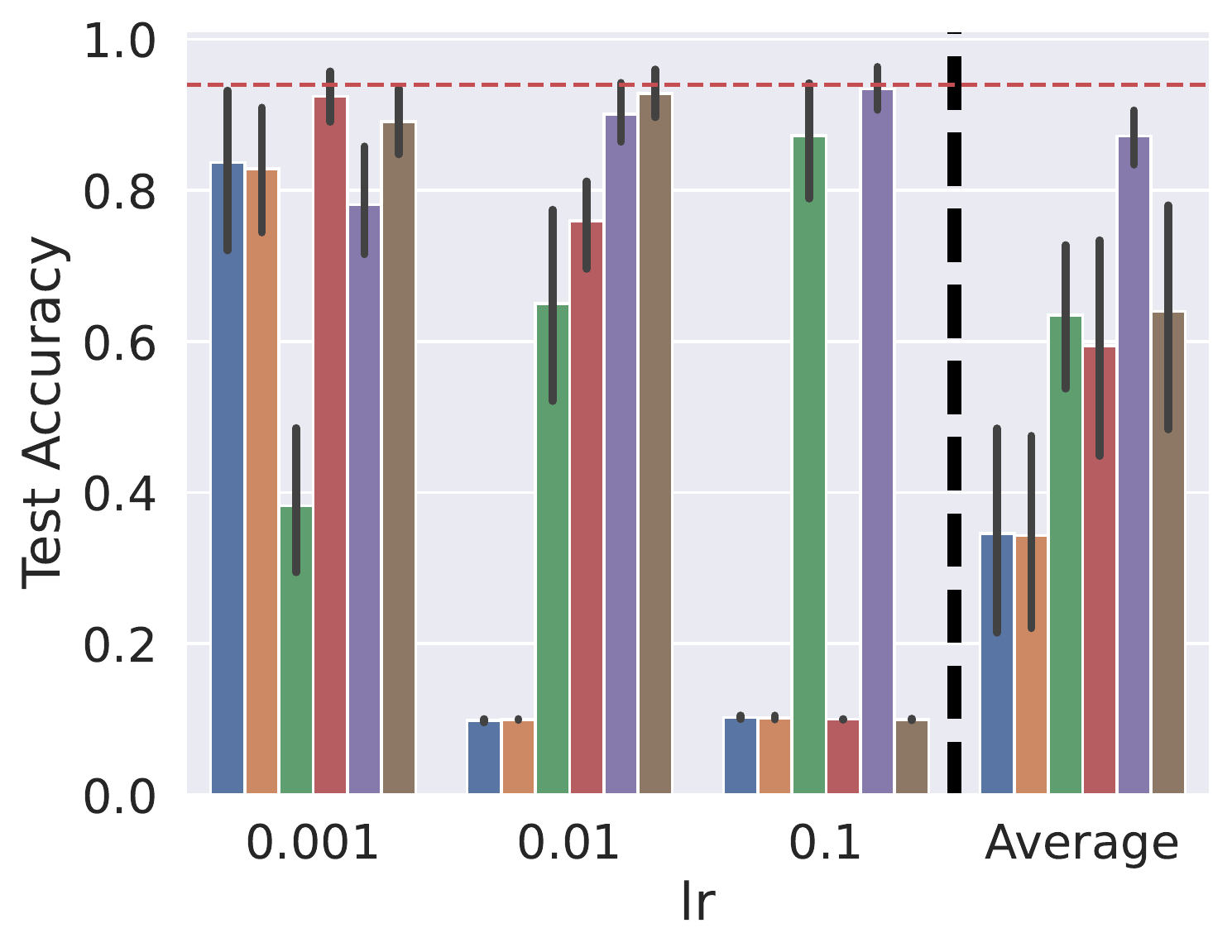}
    \caption{HPs search}
    \label{fig:preliminary_lr}
    \end{subfigure}
    \caption{Knowledge accumulation on MNIST over a short sequence of tasks (a), when scaling the number of tasks (b). Estimation of forgetting through time (c). The Hyper-parameters search (d) is realized by averaging performance on scenarios built from MNIST, Fashion MNIST and KMNIST. Results show that masking gradient and removing momentum on SGD consistently increase knowledge accumulation leading to a reduction of forgetting.}
    \label{fig:first}
\end{figure}
In this first experiment, we evaluate knowledge accumulation in \Scole{} using MNIST dataset. We use the two most popular optimizers in CL and ML in general: SGD with momentum~\citep{qian1999momentum}, and Adam~\citep{kingma2014adam}. By default, we kept the optimizer's hyper-parameters of PyTorch~\citep{NEURIPS2019_9015} and train a small convolutional neural network (c.f. \cref{ap:sec:cnn}).

Inspired by recent papers that show that masking the gradient in the last layer helps classifiers to learn continually~\citep{caccia2022new,zeno2018task} even without any supplementary memorization process~\citep{lesort2021continual}.  
We also propose to use two simple modifications to SGD (1) removing the momentum, (2) masking the output layer's gradient for classes not currently in the task in the output layer: \say{gradient masking}. More precisely, for masking, we replace outputs of classes that are not present in the current mini-batch by $-1e9$ to prevent backpropagation and to have minimal influence on the softmax activation.
%
%
On the one hand, intuitively, in the presence of data distribution drifts, momentum (also present in Adam) produces a mixture between the gradient of the previous task and the gradient of the current task, which can create interference in the training process.
On the other hand, when training on a subset of classes, the model does not learn anything about other classes. Hence masking the gradient for those outputs avoids interfering with updates for other classes.

 We see in \cref{fig:first_imp} that during the first tasks, no baseline seems to accumulate knowledge, which is in accordance with the common understanding of the CF phenomenon: CF erases past knowledge, and the overall performance depends only on the current task~\citep{French99}.   
However, when we scale the number of tasks and let reoccurrences happen, in \cref{fig:first_scaling}, we see that performance increases quickly  until reaching IID accuracy, at least for most baselines. 
We also see that forgetting decreases through time in \cref{fig:first_imp_fg}.
We note that since classes reoccur, forgetting needs to be measured differently than in usual scenarios \citep{van2019three} where we track accuracy on all tasks seen so far.
We propose to evaluate forgetting within a task by measuring the decrease in accuracy among classes seen so far and not in the current task. This gives us the average performance lost in known classes while learning the current task. The details of the calculation are in \cref{ap:sec:fg_details}.

\subsection{Robustness of Knowledge Accumulation}

\textbf{Baseline and Hyperparameters:} To find the best setup to run our further experiments, we run a small hyperparameter search on the same scenario with MNIST, Fashion-MNIST, and KMNIST. 
\cref{fig:preliminary_lr} shows the average performance on the three datasets with various learning rates. SGD without momentum and with masking is a stable baseline, as it leads to knowledge accumulation consistently with all learning rates.  Hence, We will experiment with this baseline in further experiments. 
Adam and SGD with momentum can also achieve knowledge accumulation, even without masking; however, our results show it is more sensitive to HPs.

\textbf{Datasets and Architectures:} We investigate if knowledge accumulation happens consistently among datasets and architectures.
\begin{wrapfigure}[19]{r}{0.5\textwidth}
    \centering
    \begin{subfigure}[b]{0.49\linewidth}
    \centering
    \includegraphics[width=\linewidth]{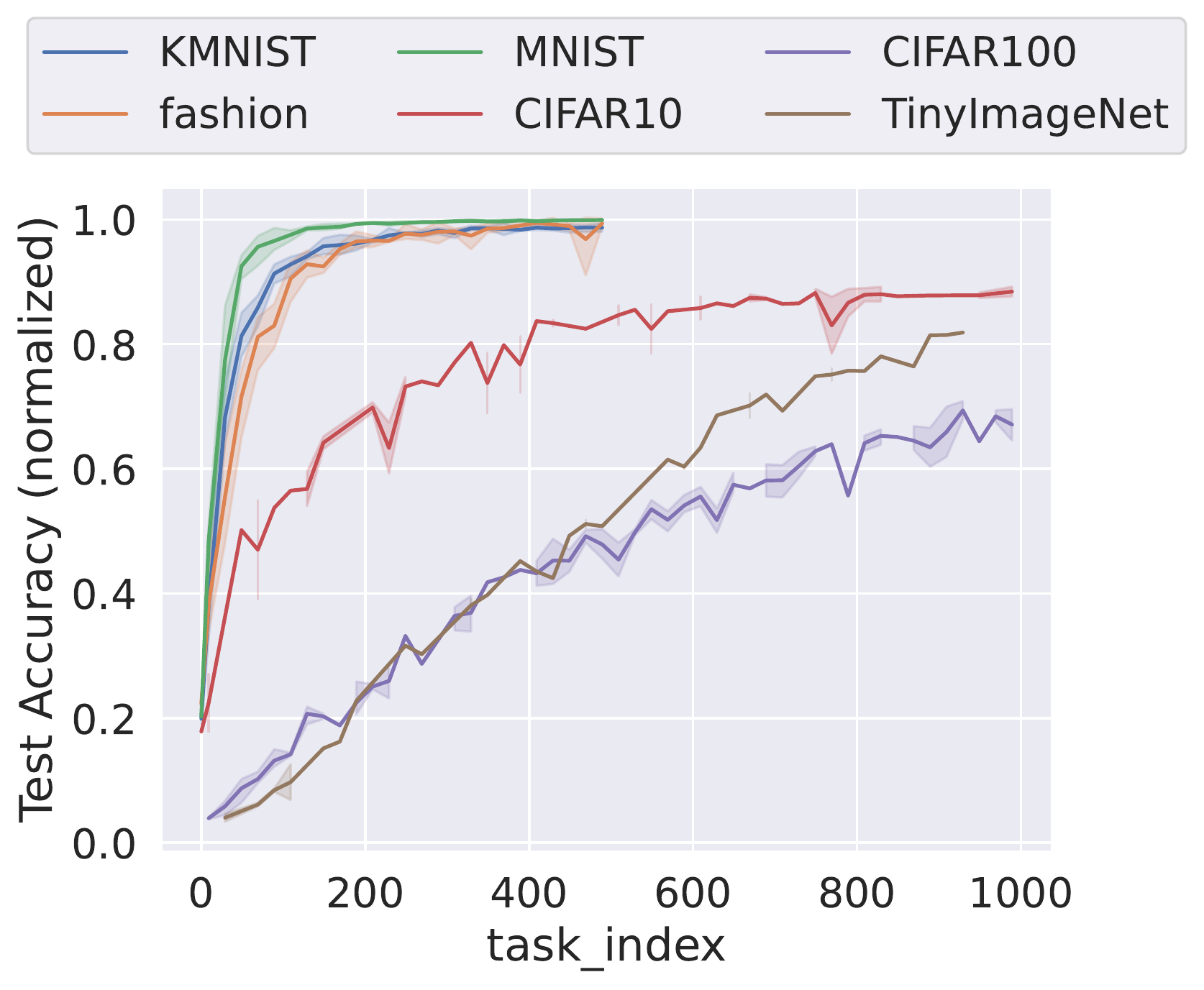}
    \caption{Datasets}
    \label{fig:preliminary_learning_curve}
    \end{subfigure}
    \begin{subfigure}[b]{0.49\linewidth}
    \centering             
    \includegraphics[width=\linewidth]{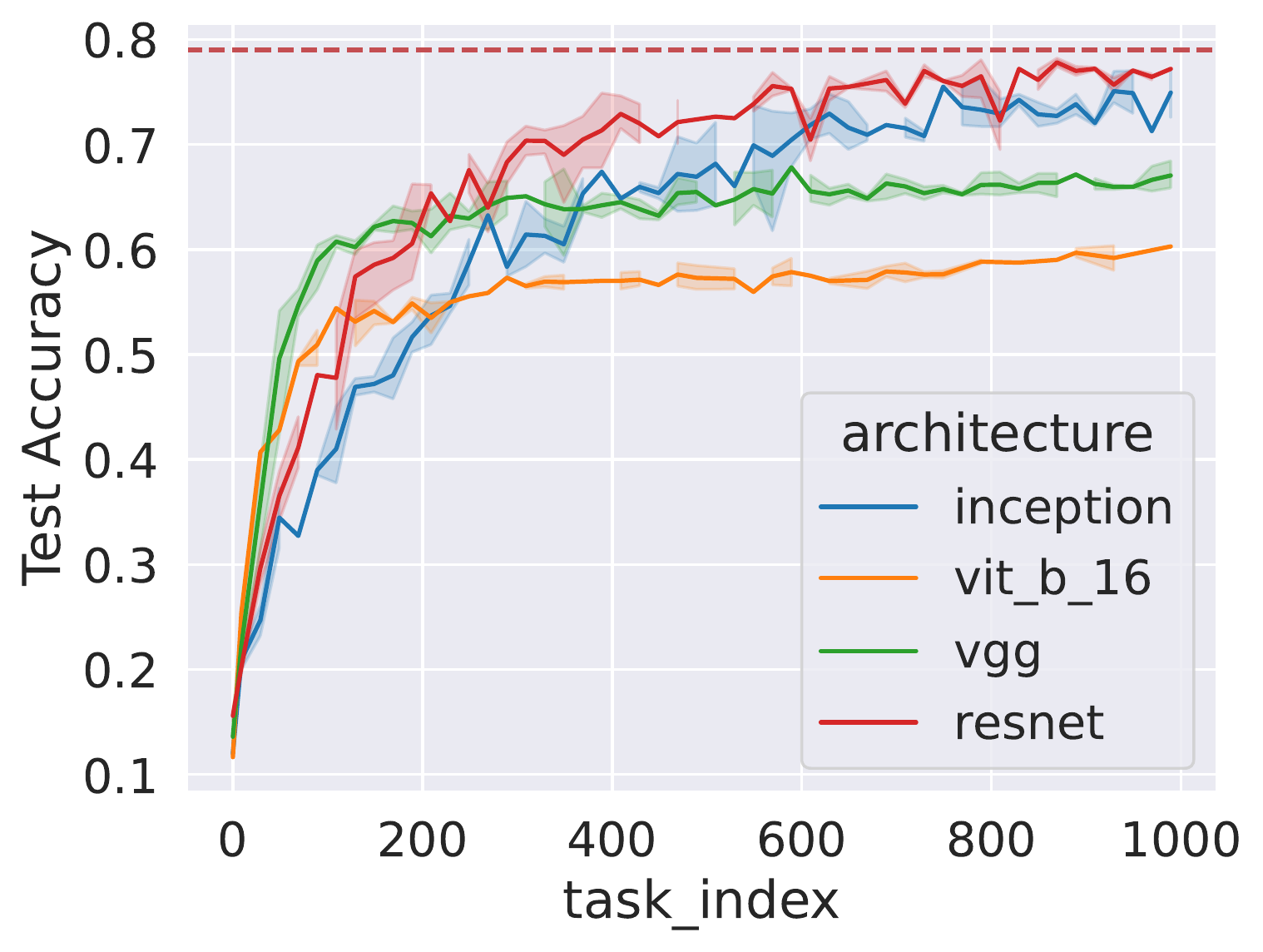}
    \caption{Architectures}
    \label{fig:architectures}
    \end{subfigure}
    
    \caption{ (left) Test acc. on \Scole{} with $T$ scaled up to 1,000 tasks averaged over MNIST, Fashion-MNIST, KMNIST with 3 seeds. 
    (left) normalized test accuracy with various datasets on Scole (the accuracy is divided by the IID accuracy),  and (right) experiment with various architectures on CIFAR10, (2 classes per tasks). 
    \protect\tikz[baseline]{\protect\draw[red,line width=0.2mm, dashed] (0,.5ex)--++(0.5,0) ;} line is the best IID performance. 
    The learning rate is set at $0.01$. }
\end{wrapfigure}
We create scenarios on MNIST, Fashion-MNIST and KMNIST with 500 tasks, CIFAR10, CIFAR100 and tinyImageNet with 1,000 tasks (3 seeds). We experiment with the baseline SGD+Mask on all of them. For CIFAR100 and tinyImageNet, we set $C=5$, while for the other datasets $C=2$.
In addition, we train several architectures (Resnet18, Inception, vit\_b\_16 and VGG) from the torch library and compare them in a default \Scole{} scenario on CIFAR10 with 2 classes per task.
 \cref{fig:preliminary_learning_curve} shows the learning curve on the various datasets. We normalize accuracies by the IID accuracy to make curves comparable.
The IID test accuracies are: MNIST $99\%$, Fashion MNIST $89\%$, KMNIST $94\%$, CIFAR10 $79\%$, CIFAR100 $40\%$, and miniImageNet $20\%$. The IID accuracies were obtained with the same models as in \Scole{} experiments, with Adam with default parameters and without data augmentation.
This figure shows that knowledge accumulation occurs consistently in all these datasets.
\cref{fig:architectures} shows that on CIFAR10, knowledge accumulation consistently happens with various types of architectures.

\textbf{Summary.} In this section, we showed that when data reoccurs through long training sequences, as in \Scole{}, gradient-based optimization may accumulate knowledge and progressively improve their overall 0-shot accuracy.
This shows the limited effect that CF might have and the influence of the knowledge accumulation (measured with meta-test probing in \cref{sec:knowledge_retention}) on 0-shot performance. 

\section{Effect of non-stationarity on continual learning with SGD}
\label{sec:complexity}

In the previous section, we investigated knowledge accumulation with one frequency of occurrence per dataset and for all classes. 
In this section, we explore more in-depth the impact of varying class occurrence frequencies on the ability of DNNs to learn continually. We explore two cases, one where $P_{\Scole{}}(Y)$ is uniform; hence the expected frequency of occurrence is the same for all classes and a second where $P_{\Scole{}}(Y)$ is not uniform.

\subsection{Uniform Occurrence Frequency}
\label{sub:uniform}

\begin{wrapfigure}[17]{r}{0.5\textwidth}
    \centering
    \begin{subfigure}[b]{0.49\linewidth}
    \centering
    \includegraphics[width=\linewidth]{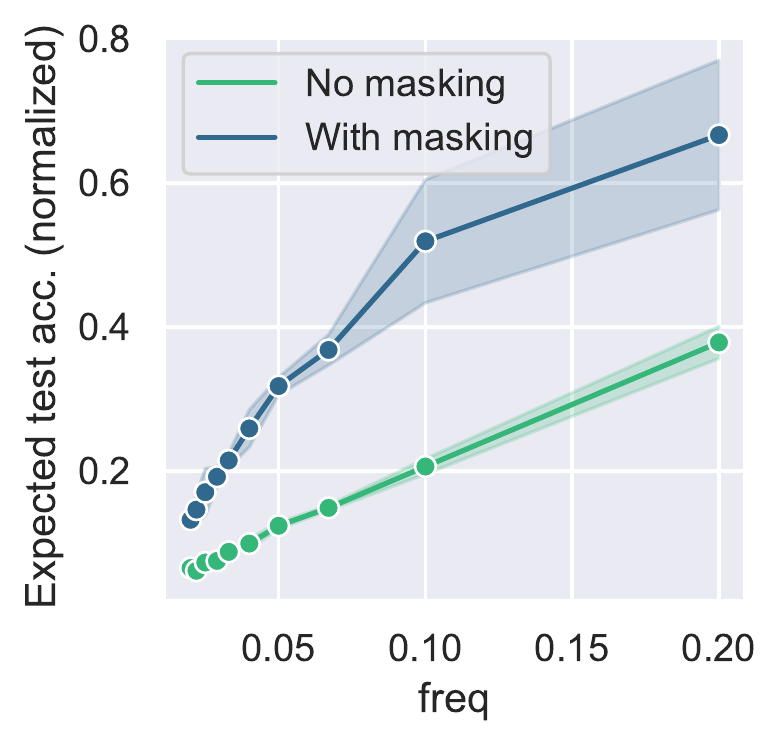}
    \caption{Occurrence Frequency}
    \label{fig:freq}
    \end{subfigure}
    \begin{subfigure}[b]{0.49\linewidth}
    \centering
    \includegraphics[width=\linewidth]{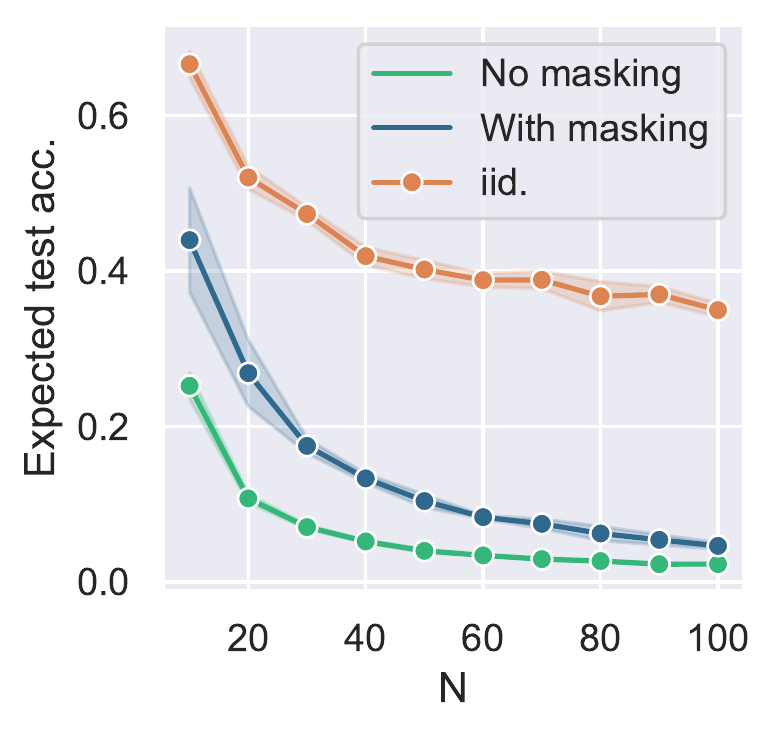}
    \caption{Number of classes}
    \label{fig:number-classes-baseline}
    \end{subfigure}        
    \caption{Expected performance depending on $\nu_{class}$ (\cref{fig:freq}). $\nu_{class}$ is influenced by the total number of classes $N$. To remove the influence of the task difficulty, we normalize by IID accuracy for each number of classes (\cref{fig:number-classes-baseline}).}
    \label{fig:number-classes}
\end{wrapfigure}

In this section, we vary the frequency of occurrence $\nu^*_{class}$ by varying the total number of classes $N$ in \Scole{} scenarios with fixed classes per tasks $C=2$ and the number of tasks $T=1000$. Growing $N$ with fixed $C$, lowers the probability of sampling each class and therefore $\nu^*_{class}$.  $P_{\Scole{}}(Y)$ is uniform, i.e.  $\nu^*_{class}$ is the same for all classes.

We use a subset of $N$ classes from CIFAR100 to create various scenarios. 
 \cref{fig:freq} shows the accuracy of scenarios with various $\nu^*_{class}$. 
 We normalized the test accuracy by the IID accuracy on the same data to only assess the effect of $\nu^*_{class}$ on the knowledge accumulation without the effect of increasing the problem difficulty. 

This experiment shows that lowering occurrence frequencies slows down knowledge accumulation.
As frequency of occurrence gets lower, further scaling of the number of tasks is necessary to get closer to IID accuracy. 
We see that for classes appearing more frequently than every 10 tasks (0.1 frequency), the expected normalized performance is higher than 50\% of the IID performance. This result demonstrates that SGD-based training can learn without forgetting and accumulate knowledge through a sequence of tasks, in particular when using masking.

\subsection{Mixture of Occurrence Frequency}
\label{sub:mixture}

\begin{figure}[!h]
    \centering
    \begin{subfigure}[b]{0.22\linewidth}
    \centering
    \includegraphics[width=\linewidth]{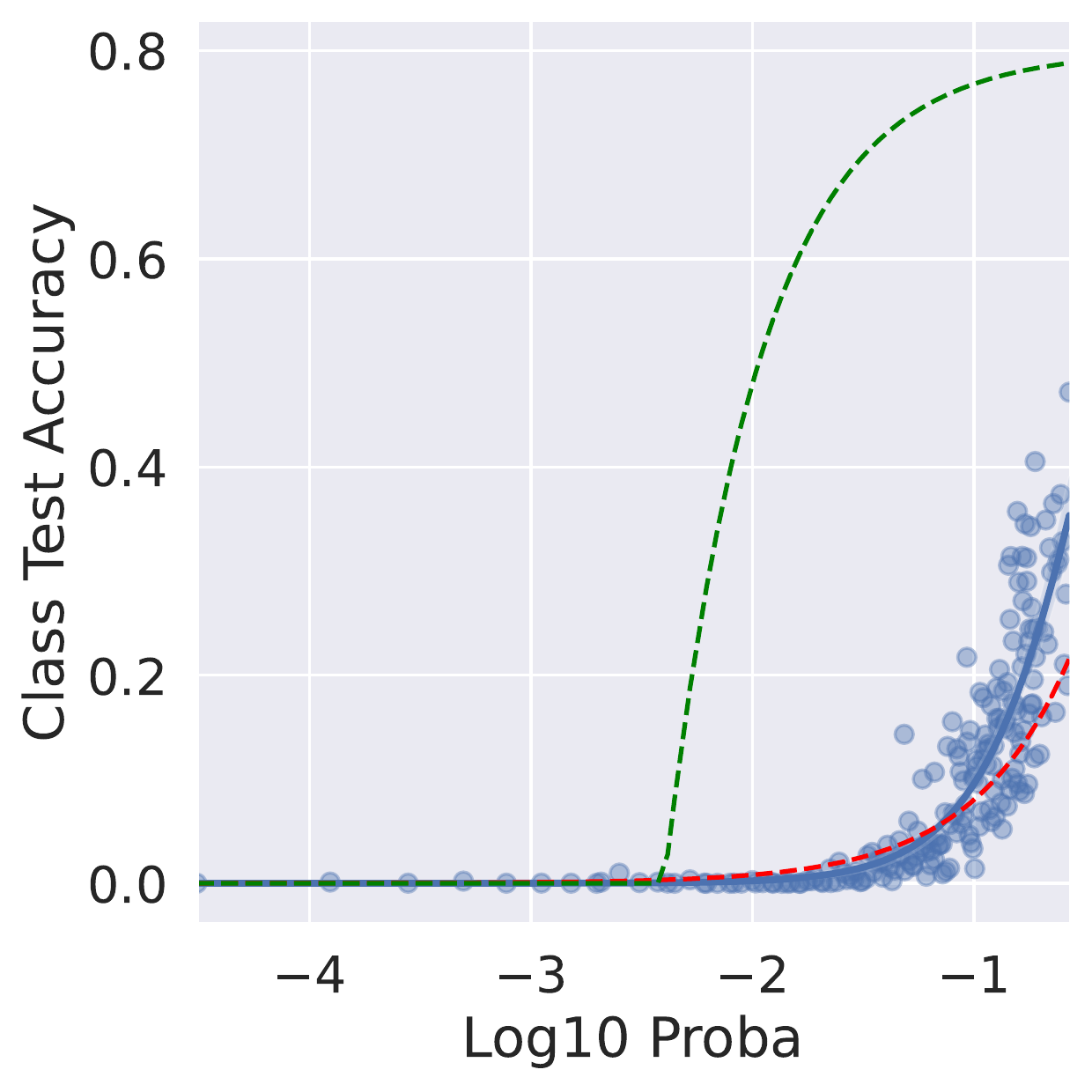}
    \caption{$t\in [0,250]$}
    \label{fig:0_250}
    \end{subfigure}
    \begin{subfigure}[b]{0.22\linewidth}
    \centering
    \includegraphics[width=\linewidth]{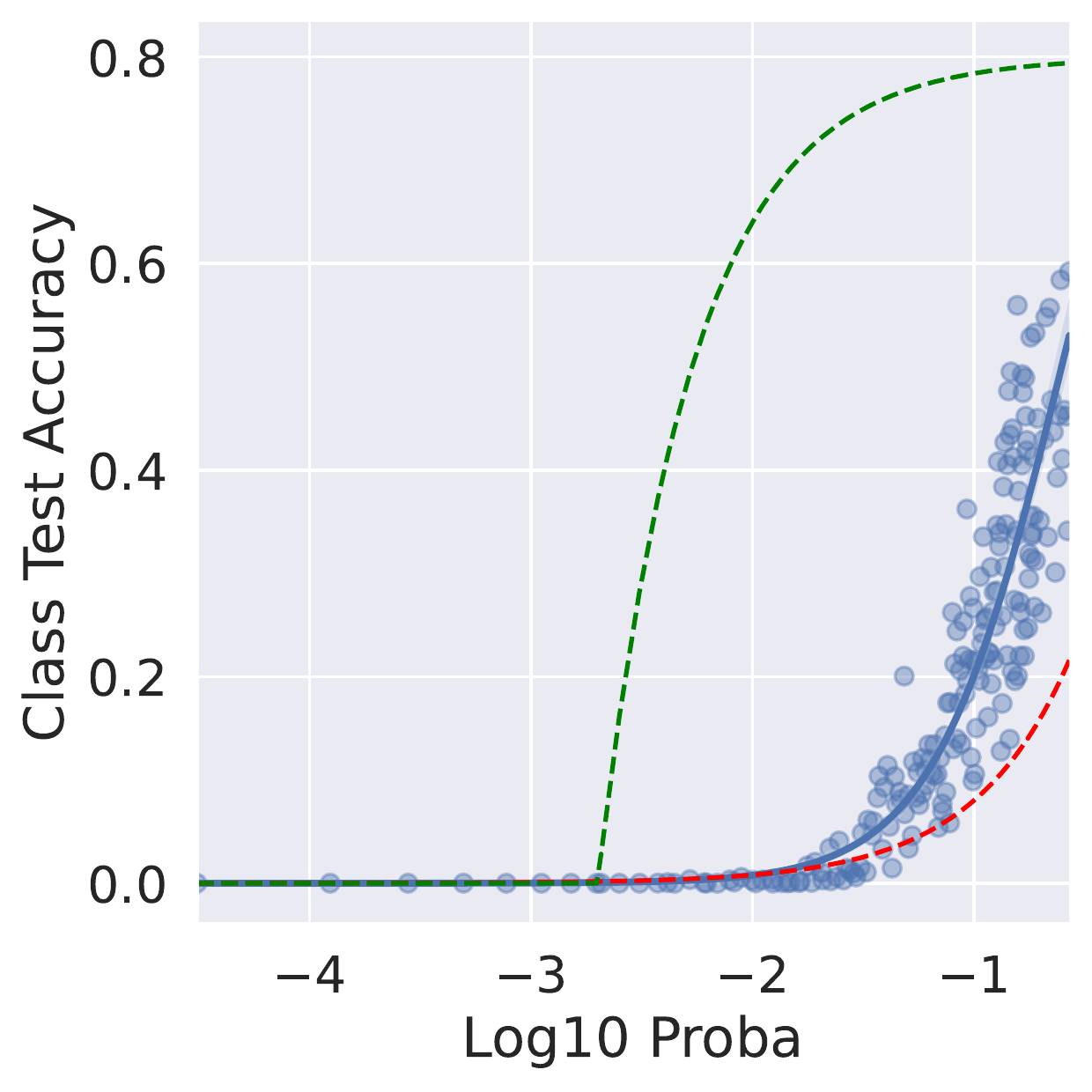}
    \caption{$t\in [250,500]$}
    \label{fig:250_500}
    \end{subfigure}
    \begin{subfigure}[b]{0.22\linewidth}
    \centering
    \includegraphics[width=\linewidth]{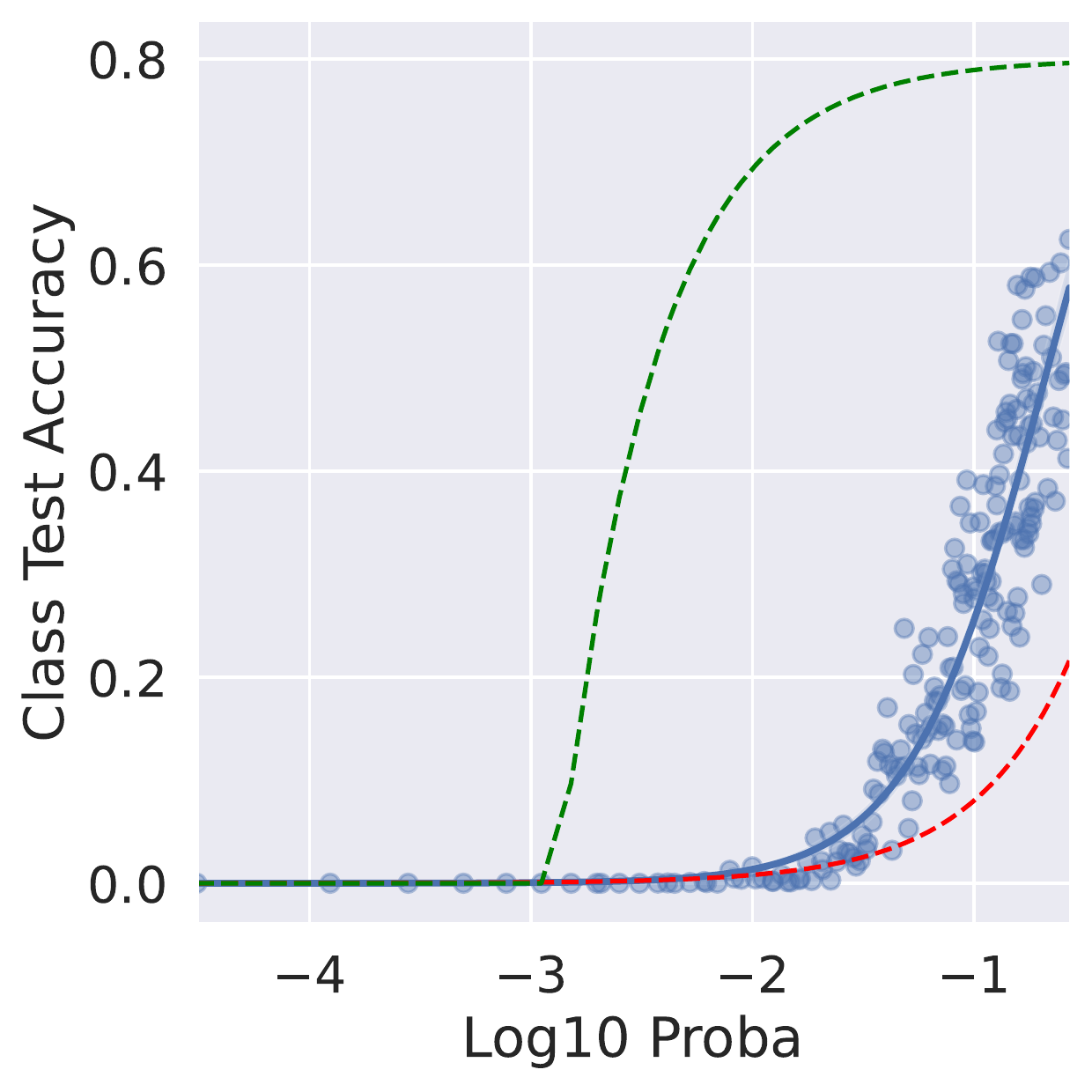}
    \caption{$t\in [500,750]$}     
    \label{fig:500_750}
    \end{subfigure}
    \begin{subfigure}[b]{0.22\linewidth}
    \centering
    \includegraphics[width=\linewidth]{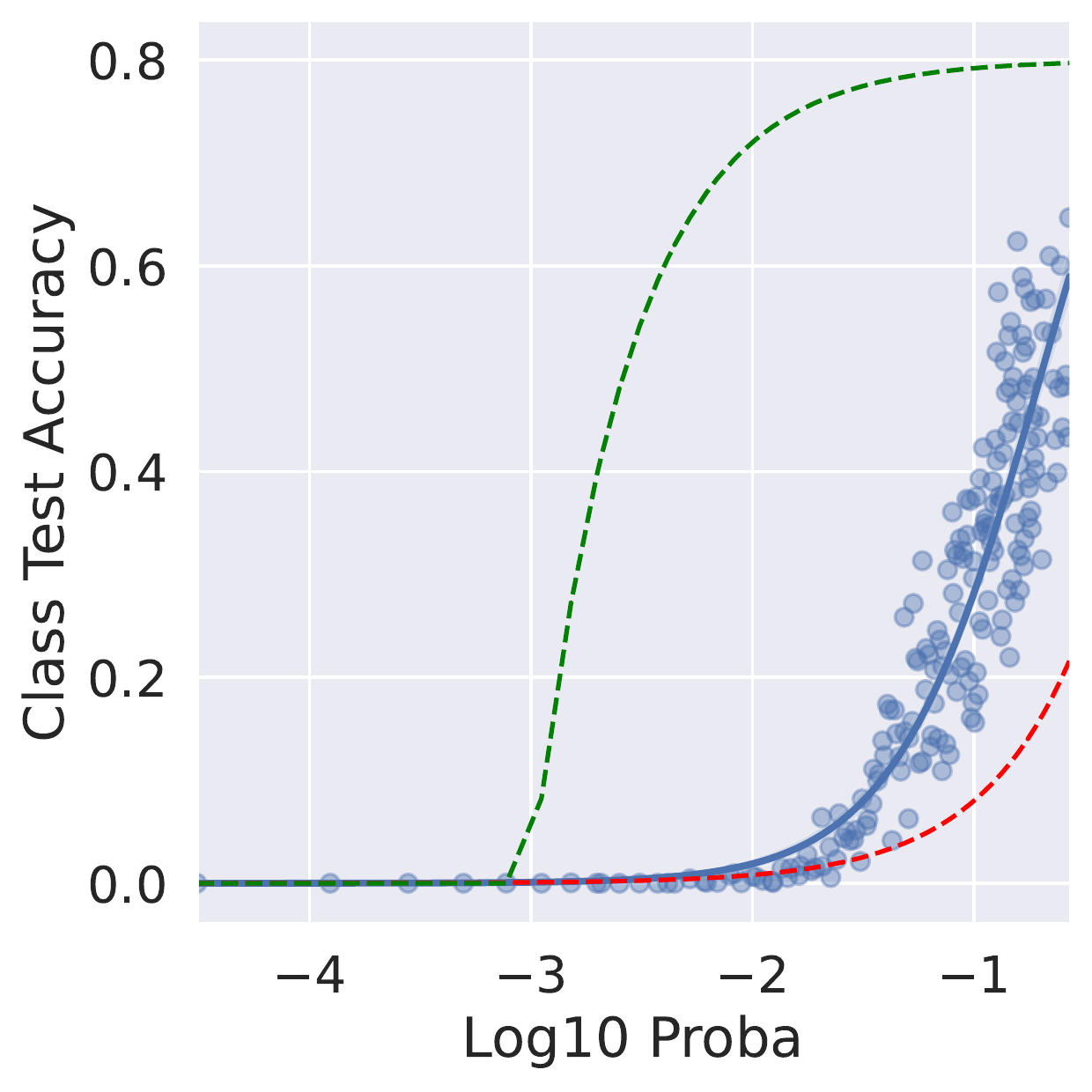}
    \caption{$t\in [750,10^3]$}
    \label{fig:750_1000}
    \end{subfigure}        
    \caption{Accuracy for classes vs occurrence probability over different task ranges. Looking at the 4 plots, we see that the model increases performance through time for classes with higher occurrence probabilities.  Lower bound \protect\tikz[baseline]{\protect\draw[red,line width=0.2mm, dashed] (0,.5ex)--++(0.5,0) ;}: Expected performance when the model learns at 80\% accuracy for current classes and forgets everything after. Upperbound \protect\tikz[baseline]{\protect\draw[green,line width=0.2mm, dashed] (0,.5ex)--++(0.5,0) ;}: Expected performance when the model learns up to 80\% each task and never forget after.}
    \label{fig:freq-mixture}
\end{figure}
    
In previous experiments, all $\nu^*_{class}$ were the same. 
In this experiment, we give a different $\nu^*_{class}$ to each class, and we analyse class performances independently from each other. This setting is built to not depend on the uniform distribution of classes in the results.

In this scenario, we use CIFAR100 dataset, $P_{\Scole{}}(Y)$ is not uniform anymore, and each class is given a probability of sampling on a wide range of probability. Then, for each task, we sample 10 classes from $P_{\Scole{}}(Y)$ to create tasks.
Details on the implementation of how to create a distribution of various probabilities are in \cref{ap:sec:entropy_decrease}. 
In this setting, we estimate $\nu^*_{class} = 1 - \Pi_{i=0}^{C-1} (1 - P_{\Scole{}}(Y=class) * \frac{N}{N-i})$ $task^{-1}$.

We can observe in \cref{fig:freq-mixture} the performance at different task intervals. Although one might expect to observe improvement across all frequency ranges, the results indicate that only the most frequent classes display enhanced performance. This finding suggests that to improve results on comparable benchmarks, it may be advantageous to prioritize low-frequency classes. Additionally, while the model outperforms the complete forgetting baseline, it still lags significantly behind a model that never forgets.

\textbf{Summary.} In this section, we empirically evaluate how SGD with masking accumulates knowledge with various class occurrence frequencies. The evaluation was conducted both for equal (uniform $P_{\Scole{}}(Y)$) and distinct class frequencies. Our experiments detail how well classes are learned and remembered depending on their occurrence frequency. One of our results is that after a significant number of tasks and when using masking, CIFAR100 classes reoccurring more than once every 10 tasks are remembered with over $40\%$ of accuracy. 

\section{Increasing Knowledge Accumulation in DNNs}
\label{sec:increasing}

\subsection{Learning Forgetting Trade-off}
\label{sec:epoch-per-task}

\begin{figure}
\centering
    \begin{subfigure}[b]{0.49\linewidth}
    \centering
    \includegraphics[width=0.7\linewidth]{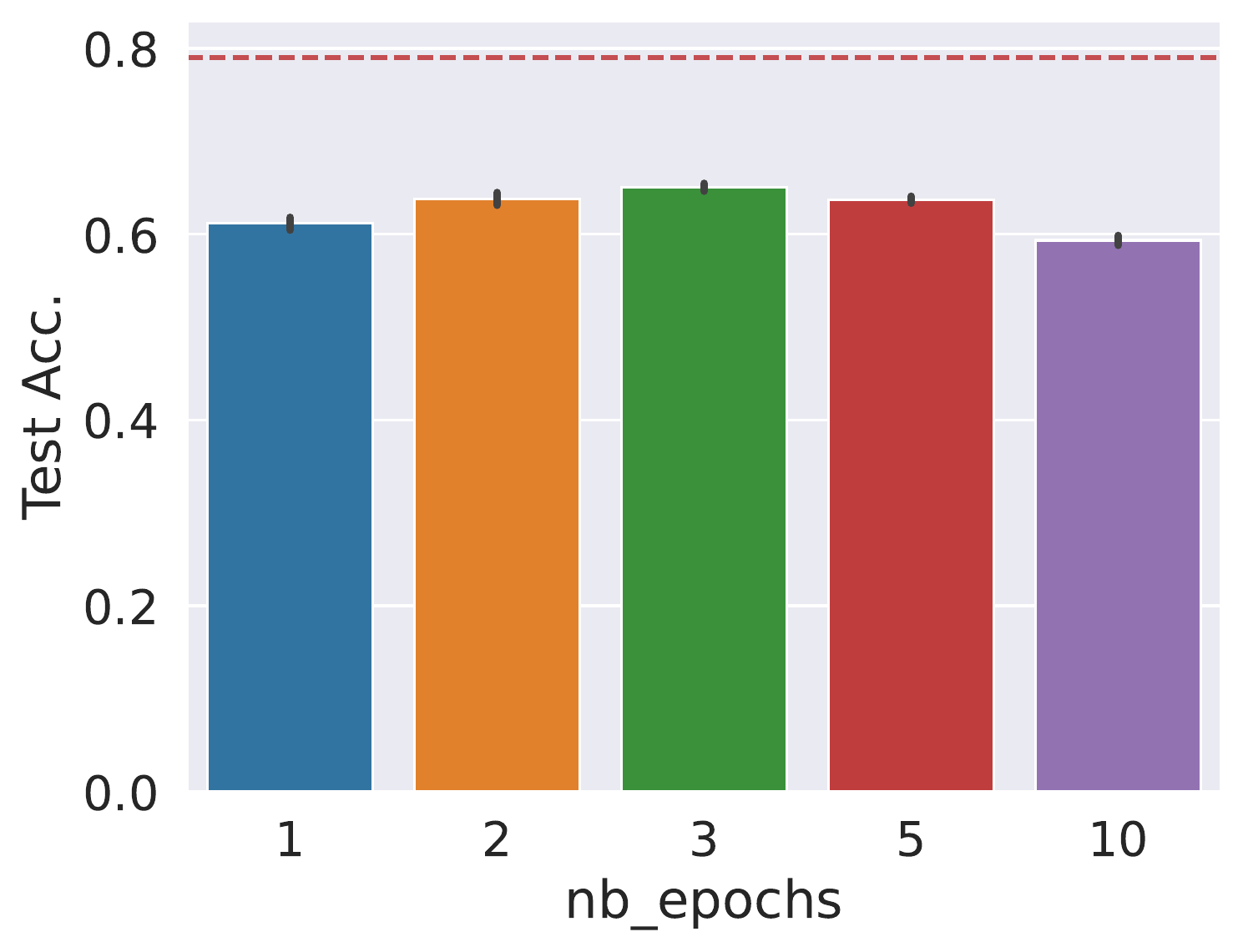}
    \end{subfigure}
\centering
    \begin{subfigure}[b]{0.49\linewidth}
    \centering
    \includegraphics[width=0.7\linewidth]{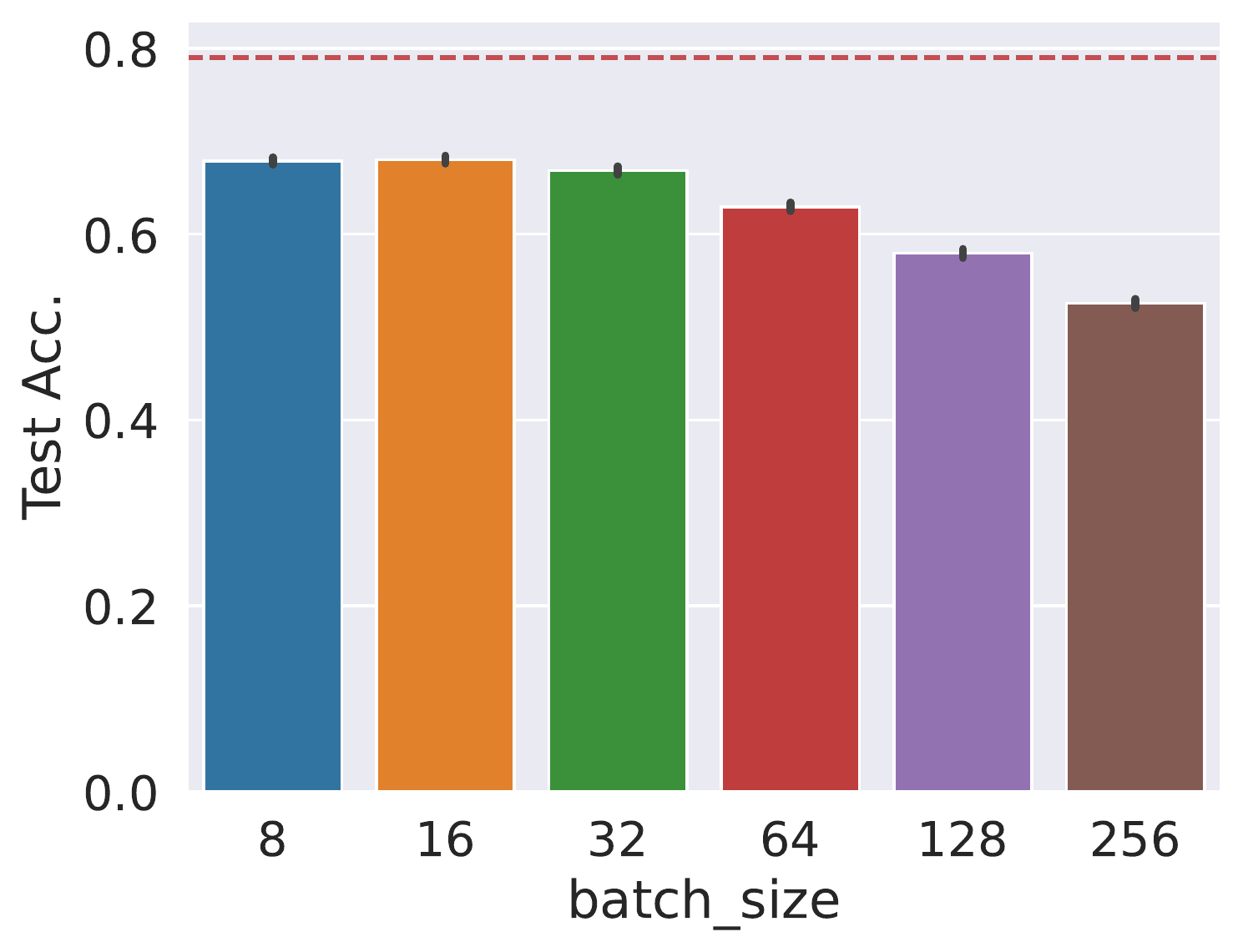}
    \end{subfigure}
    \caption{Average test acc. on 1k tasks vs epochs per task (top) and batch size (bottom). Performance may improve with more gradient steps per task but too many gradient steps on each task might lead to performance loss. \protect\tikz[baseline]{\protect\draw[red,line width=0.2mm, dashed] (0,.5ex)--++(0.5,0) ;} IID baseline.}
    \label{fig:learning-forgetting}
\end{figure} 

A straightforward approach to minimize the effect of catastrophic forgetting could be to minimize the number of training steps in each task. It would minimise forgetting during a given task, maximize the randomization of data, and get closer to an IID training regime. On the other hand, it could also significantly limit the capability to learn the tasks themselves.

In this section, we investigate this strategy. We introduce the concept of \say{learning-forgetting trade-off}. While learning on a new task, the model increases its knowledge about this new data but also might forget what was learned before. The right learning-forgetting trade-off maximizes overall knowledge and aims to ensure that learning brings more knowledge than forgetting.

\textbf{Setting:} In this experiment, we use the CIFAR10 dataset with $C=2$ classes per task uniformly sampled with a scenario of $N=1000$ tasks. We vary the batch size and the number of epochs to control the number of gradient steps per task. 

\textbf{Results:} \cref{fig:learning-forgetting} shows that the number of epochs per task or the batch size that maximize the performance are not the ones that minimize the number of gradient steps per task. Growing the number of steps per task might actually improve performance, i.e. the model learns more than it forgets, at least until a certain point, as shown when we increase the number of epochs. 
If the gradient steps' number is increased too much, the model starts to forget more than it learns. \modif{We can note that the ordering of the best number of epochs or best batch size might evolve with the number of tasks. For more detail, we present the result for the same experiments with the test accuracy per task in appendix \cref{ap:fig:learning-forgetting}.}

This experiment shows the trade-off between learning new information and forgetting previous ones. At scale, algorithms should then find the right amount of new information to learn to maximize this trade-off and the overall performance.

\subsection{Model Width}
\label{sec:model-width}

\begin{wraptable}[12]{r}{0.6\linewidth}
\caption{Average Performance at different $\nu_{class}$ range over the last 100 tasks (over 1000) of various model widths.}
\label{tab:width}
\centering
\begin{tabular}{|p{2cm}|p{1.2cm}|p{1.2cm}|p{1.2cm}|p{1.2cm}|}

\toprule
\tiny{\textbf{Size $\backslash$ Probabilities}} & 
\tiny{$10^{-4} \leq p<10^{-3}$} &
\tiny{$10^{-3} \leq p<10^{-2}$} &
\tiny{$10^{-2} \leq p<10^{-1}$} &
\tiny{$10^{-1} \leq p<1$} \\ \midrule

$1$ & 
\tiny{0.0 \%} & 
\tiny{0.2 \%} & 
\tiny{15.4 \%} &
\tiny{41.4 \%} \\

$2$ & 
\tiny{0.0 \%} & 
\tiny{0.2 \%} & 
\tiny{18.2 \%} &
\tiny{43.8 \%} \\
 
$4$ & 
\tiny{0.0 \%} & 
\tiny{0.4 \%} & 
\tiny{22.3 \%} &
\tiny{44.2 \%} \\
 
 $8$ & 
\tiny{0.0 \%} & 
\tiny{0.7 \%} & 
\tiny{26.4 \%} &
\tiny{44.5 \%} \\

 $16$ & 
\tiny{0.0 \%} & 
\tiny{1.6 \%} & 
\tiny{32.0 \%} &
\tiny{45.9 \%} \\
 
 \bottomrule
\end{tabular}
\end{wraptable}


\modif{Inspired by \cite{mirzadeh2021wide}, in this section, we would like to investigate how different widths of resnet models learn on \Scole{} setup. We are also interested in understanding if making models wider improve performance in all range of frequency of appearance.}

\modif{We define the bandwidth of SGD as the $\nu_{class}$ range that SGD is able to learn without forgetting and accumulating knowledge. 
We train on the CIFAR100 dataset with SGD (lr=0.1, momentum=0) on one epoch per task. All classes have a different probability of being sampled when building a task as in \cref{sub:mixture}. } 
%

\modif{We present the results of this experiment on various $\nu^*$ ranges in \cref{tab:width}.  As in \cite{mirzadeh2021wide}, the \Scole{} setup confirms that wider models perform better. Moreover, interestingly, we see that most of the performance improvement is focused on the intermediate frequency range of $\nu_{class} \in [1\mathrm{e}{-2},1\mathrm{e}{-3}]$. }

\modif{These results show that wider models improve performance but also increase the bandwidth of SGD with masking.}

\subsection{Frequency Replay}
\label{sub:replay}

As we have seen in previous sections, SGD training on DNNs leads to knowledge accumulation, a fortiori if the occurrence frequency of classes is high. 
In this section, we show how to benefit from a better understanding of knowledge accumulation to improve the compute efficiency of a vanilla replay approach.

\textbf{Approach: } 
Replaying past data is a standard approach to CL. However, replay approaches usually do not take into account the capability of SGD to accumulate knowledge.
Replay strategies are either based on random sampling \citep{rebuffi2017icarl,Chaudhry2019Continual} or based on current loss space \citep{borsos2020coresets,aljundi2019online}.
Therefore, as a proof of concept that we can benefit from both knowledge accumulation and replay mechanisms, we propose to replay based only on the class occurrence frequencies. The data from all classes is saved in the same way, but data from classes that appear frequently are not replayed.

Based on experiments in \cref{sub:mixture} and \cref{fig:freq-mixture}, we assume that our model effectively sufficiently accumulates knowledge for classes with $\nu_{class} > 0.1~tasks^{-1}$ and consequently, we replay other classes. However, to limit the compute cost of our replay mechanisms, we do not replay too rare classes ($\nu_{class} < 0.01~tasks^{-1}$). The \textit{passband} of our frequency replay mechanism, i.e. the $\nu_{class}$ range that is replayed, is $[0.01,0.1]$.
To resume, the frequency replay mechanisms target classes that appear every 10 to 100 tasks.

\begin{wrapfigure}[21]{r}{0.5\textwidth}
    \centering
    \begin{subfigure}[b]{0.49\linewidth}
    \centering
    \includegraphics[width=\linewidth]{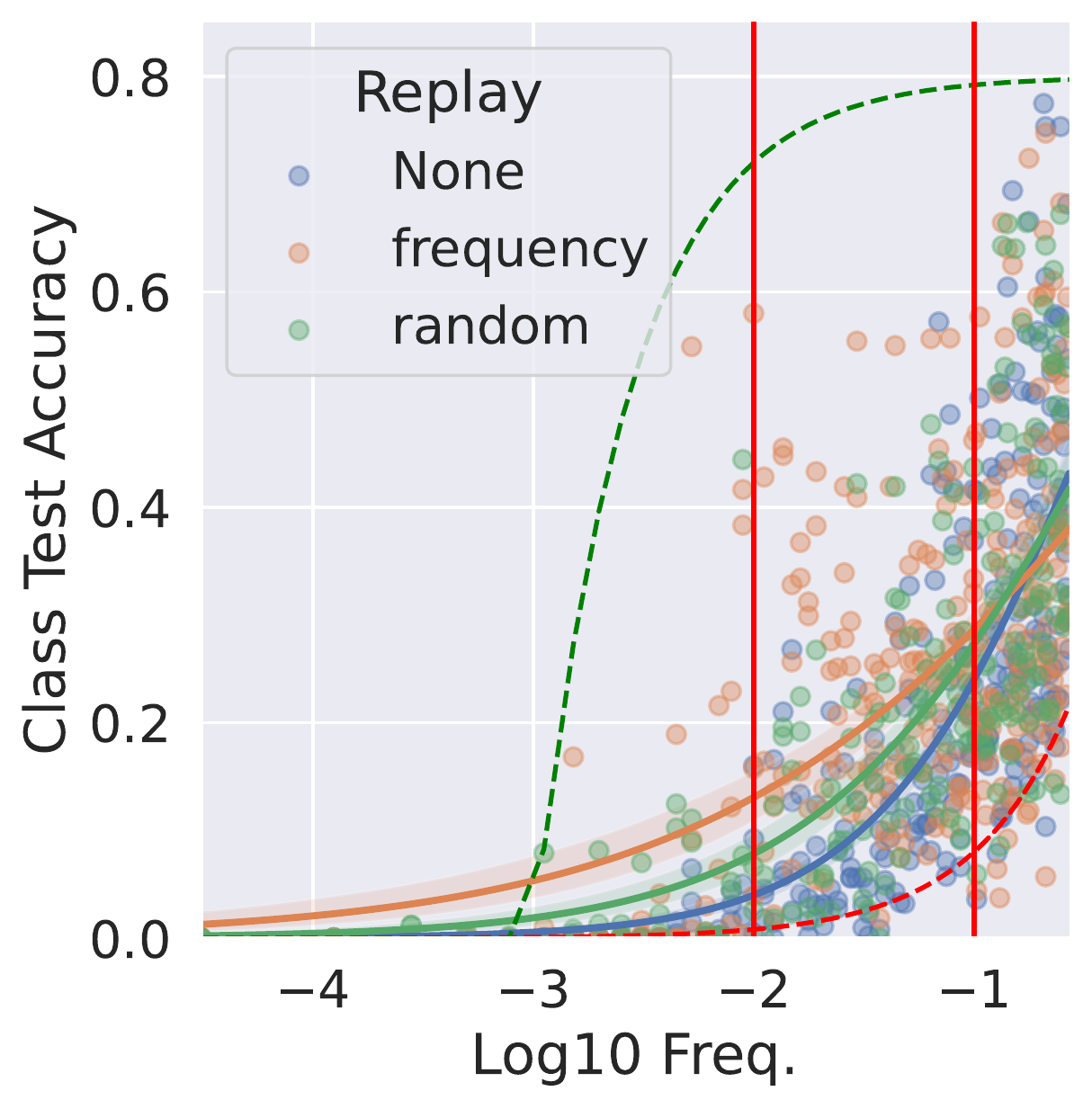}
    \caption{Frequency Replay.}
    \label{fig:frequency-replay}
    \end{subfigure} 
    \begin{subfigure}[b]{0.49\linewidth}
    \centering
    \includegraphics[width=\linewidth]{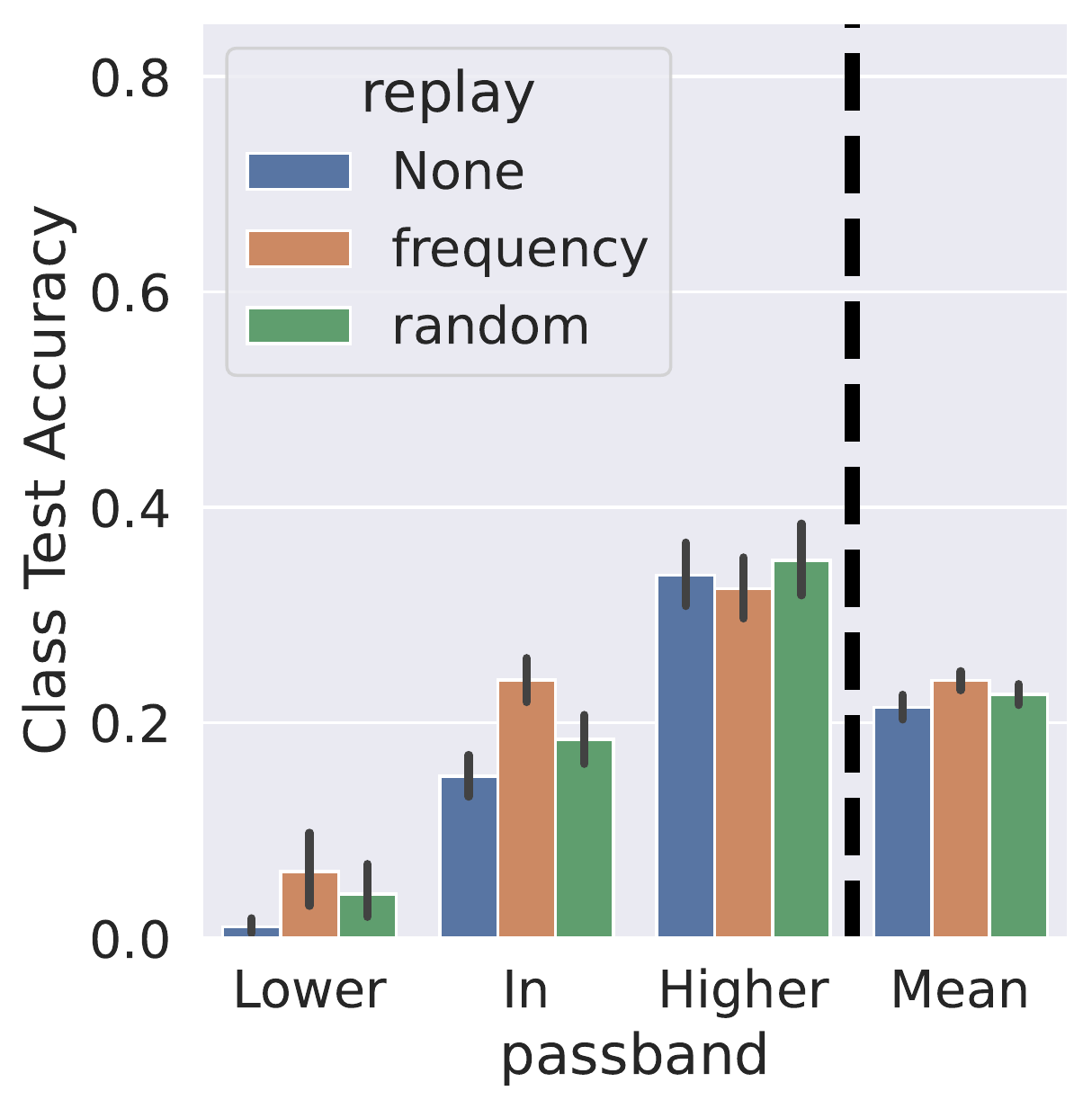}
    \caption{Acc. per frequency range}
    \label{fig:passband}
    \end{subfigure} 
    \caption{Class mean performance over the last 100 tasks vs $\nu^*_{class}$ (log). In \cref{fig:frequency-replay}, the vertical red lines are the upper and lower cutoff frequencies of the replay mechanism. \cref{fig:passband} plots the average acc. depending on if $\nu^*_{class}$ is within the passband of the replay mechanism, lower or higher. Frequency replay improves performance in the desired frequency range and on average.}
    \label{fig:freq-replay}
\end{wrapfigure} 

\textbf{Implementation Details: } Each $\nu_{class}$ is estimated by counting occurrences during the scenario. 
To have an estimate of frequency on several occurrences of one class, the replay process may be triggered after $\tau=3$ occurrences (c.f. \cref{alg:freq-replay}).
The buffer saves $n=200$ samples per class. For each revisit, $max(n/o, n/20)$ samples are renewed, where $o$ is the number of occurrences of the class so far. We set a minimum of renewable by principle to avoid the replay buffer becoming frozen.
While replaying, we oversample the buffer in order to keep a number of samples per class balanced during the task as in \citet{Ostapenko2022Foundational}.

In this setting, we evaluate the advantage of the frequency replay strategy. We compare frequency replay with simple fine-tuning and random replay. To increase the difficulty of the previous settings and avoid the repetition of the exact same data point, we use TinyImagenet dataset with the augmentations proposed in \citet{hendrycks2019benchmarking}. Those augmentations simulate common perturbation and corruption in images. At each task, a random augmentation is selected and applied with severity 2. 
We use a subset of 100 classes of TinyImagenet and train with 10 classes per task to have the same probability distribution as in \cref{sub:mixture}.
In this setting, the compute overload of frequency replay over normal fine-tuning is $21.28\%$. In comparison, doing a full replay strategy, i.e. replaying all classes seen so far (and not in the current task), would increase the compute by $873\%$ (details in \cref{ap:sec:compute}). 
%
In our experiments, instead of comparing with full replay, we compare the frequency replay with random replay with the same compute budget. \modif{The experiments are run on 10 different seeds.}

  \begin{wrapfigure}[19]{r}{0.4\textwidth}
  \begin{minipage}{0.4\textwidth}
\begin{algorithm}[H]
   \caption{Frequency Replay}
   \label{alg:freq-replay}
    \begin{algorithmic}
       \STATE {\bfseries Input:} $\nu_{low}$, $\nu_{high}$, $\tau$, $C_{dict}$, $C_{t}$, nb\_batch.
       \STATE $\mbox{nb\_batch} \gets \mbox{nb\_batch}+1$
       \STATE $classes2replay=[]$
    
        \FOR{$class=0$ {\bfseries in} $C_{t}$}
        \STATE $C_{dict}[class]+=1$
       \ENDFOR
       
       \FOR{$class=0$ {\bfseries in} $C_{dict}.keys() \backslash C_{t}$}
       \STATE $freq = C_{t}[class] / \mbox{nb\_batch}$
       \IF{$C_{t}[class] > \tau$ \& $freq \in [\nu_{low}, \nu_{high}]$}
       \STATE classes2replay.append(class)
        \STATE $C_{dict}[class]+=1$
       
       \ENDIF
       \ENDFOR
       \STATE {\bfseries Return:} classes2replay
    \end{algorithmic}
    \end{algorithm}
    \end{minipage}
\end{wrapfigure}

We gathered results in \cref{fig:freq-replay}. This figure shows the positive effect of frequency replay within the frequency range of interest and overall (Average) compared to fine-tuning and random replay \footnote{\modif{The p-value of frequency replay results being greater than random replay results is $0.04$ in this experiment. With the usual significance level at 5\%, it means that the difference is significant.}}. 
We observe that in lower frequencies, performances also improve. Indeed, the x-axis is the expected frequency of occurrence and not the empirical frequency of occurrence. If empirically, a low-frequency class appears several times at the beginning of the scenario, it will be \say{captured} by the replay mechanism and be replayed regularly.
The results with this simple replay mechanism illustrate as a proof of concept the impact of better understanding knowledge accumulation in DNNs to improve the scalability and efficiency of continual learning approaches.
\modif{A related idea to our frequency replay, was proposed in \cite{Hemati2023ClassIncrementalLW}, where the amount of data stored in the replay buffer depends on the frequency of appearance of classes. However, their approach is meant to optimize memory footprint while our is meant to optimize compute footprint.}

\textbf{Note:} In classical CL scenarios, classes do not reoccur throughout the sequence of tasks. Nevertheless, frequency replay could still be used. For example, instead of replaying every 10 tasks for $\nu^*_{class}=0.1$, in a classical setup classes could be replayed every 10 epochs to reduce compute cost while maintaining the advantage of replay. \modif{This method was notably used for continual learning for text classification and question answering in \cite{Autume2019Episodic}.}

\textbf{Summary.} Making models wider, replaying based on frequency and/or optimizing the learning-forgetting trade-off strategies are scalable and efficient approaches to improve the knowledge accumulation of SGD. 
Those results motivate the importance of understanding the dynamics of learning and forgetting in the long run in order to design algorithms that achieve optimal performance-compute trade-offs in practical settings.

\section{Related Work}
\label{sec:related_work}

Most scenarios in the continual learning literature study catastrophic forgetting \citep{van2019three,lesort2021understanding}. Such scenarios consist of a sequence of tasks where data appears in a single task. Those settings evaluate whether models can remember tasks that they have seen only once.  
The no reappearance constraint makes the evaluation of models clear since tasks overlap can not interfere with forgetting. However, they cannot study if approaches such as SGD retain and accumulate knowledge through time. This is the main difference with the present work. 

The scenario we propose has some similarities to the CMR \citep{lin2022continual} and OSAKA \citep{caccia2020online} frameworks. However, in our setting, we do not evaluate fast adaptation but rather the capacity to learn classes and remember them depending on their occurrence frequency. Moreover, there is no real concept shift in \Scole{}. That is, $p(y|x)$ is fixed over time.
%
%
A scenario with a long sequence of tasks was proposed by \citet{Wortsman2020Supermasks}. Different permutations of pixels for each task are used. However, there is no task re-occurrence, and SGD cannot accumulate knowledge.
\citet{Stojanov_2019_CVPR} experimented with scenarios with the re-occurrence of objects and find out that it reduces the effect of forgetting. However, they did not analyse this finding by varying the frequency of occurrence. 

On a similar line as this work, \citet{evron2022catastrophic} theoretically showed that for linear regression trained with SGD, knowledge accumulation leads to a reduction of CF when tasks reoccur.  
Other works investigate the effect of forgetting on weights: \citet{ramasesh2021anatomy} shows that CF impact mostly the weight on the higher layers.  \citet{Doan2021Theoretical,asanuma2021statistical} use the modification of weights between two tasks as a proxy to measure the similarity of tasks.

The findings of this paper that DNNs consistently retain and accumulate knowledge could provide  an explanation for the observation that self-supervised models are continual learners \citep{fini2020online}, language models are continual learners \citep{scialom2022continual}, or that large scale pertaining reduce forgetting \citep{ramasesh2022effect}. On the one hand, in settings such as self-supervised learning and language modelling, knowledge accumulation is improved by training with optimised general losses, which increase transfer between tasks and favour the re-occurrence of concepts and features in data. On the other hand, when starting from a model pre-trained on a large task, long-term knowledge retention benefits all downstream tasks, especially in short sequences. Nevertheless, if those results might be connected into common learning and forgetting dynamics, the specificities of each setup remain a factor of variability in results.

\section{Conclusion}
\label{sec:conclusion}

This paper investigates knowledge retention (KR) and accumulation (KA) in deep neural networks. We show that contrary to what most CL literature shows, catastrophic forgetting has a limited effect on DNNs. 
We show that after learning a task, DNNs retain knowledge about it even after learning several unrelated tasks. 
Moreover, when trained on long sequences of tasks with reoccurring classes, DNNs progressively accumulate knowledge leading to an overall improvement in performance. 
%
We propose the \Scole{} framework to study this phenomenon under various occurrence frequencies of classes. As expected, the performance decreases when the frequency becomes lower; nevertheless, we find out that simple gradient-based approaches are better than expected at learning without forgetting. For example, with CIFAR100 data, SGD can reach more than $40\%$ of accuracy on data occurring once every 10 tasks.

Additionally, to improve knowledge accumulation with SGD, we experiment with simple and scalable strategies: (1) masking the gradient in the last layer, (2) optimizing the learning-forgetting trade-off, (3) making the model wider, or (4) replaying only low-frequency classes.

We believe that assessing the efficiency of algorithms with regard to frequency occurrence could greatly improve how continual learning is applied in real scenarios. Future work could evaluate existing continual learning strategies on \Scole{} to estimate their scalability to long sequences of tasks and their frequency range of efficiency.

\bibliography{continual,others}
\bibliographystyle{collas2023_conference}

\appendix

\section{Details}

\subsection{\modif{Details on Figure \ref{fig:learning-forgetting}}}
\label{ap:sec:kl}

\begin{figure}[!ht]
    \begin{subfigure}[b]{0.49\linewidth}
    \centering
    \includegraphics[width=\linewidth]{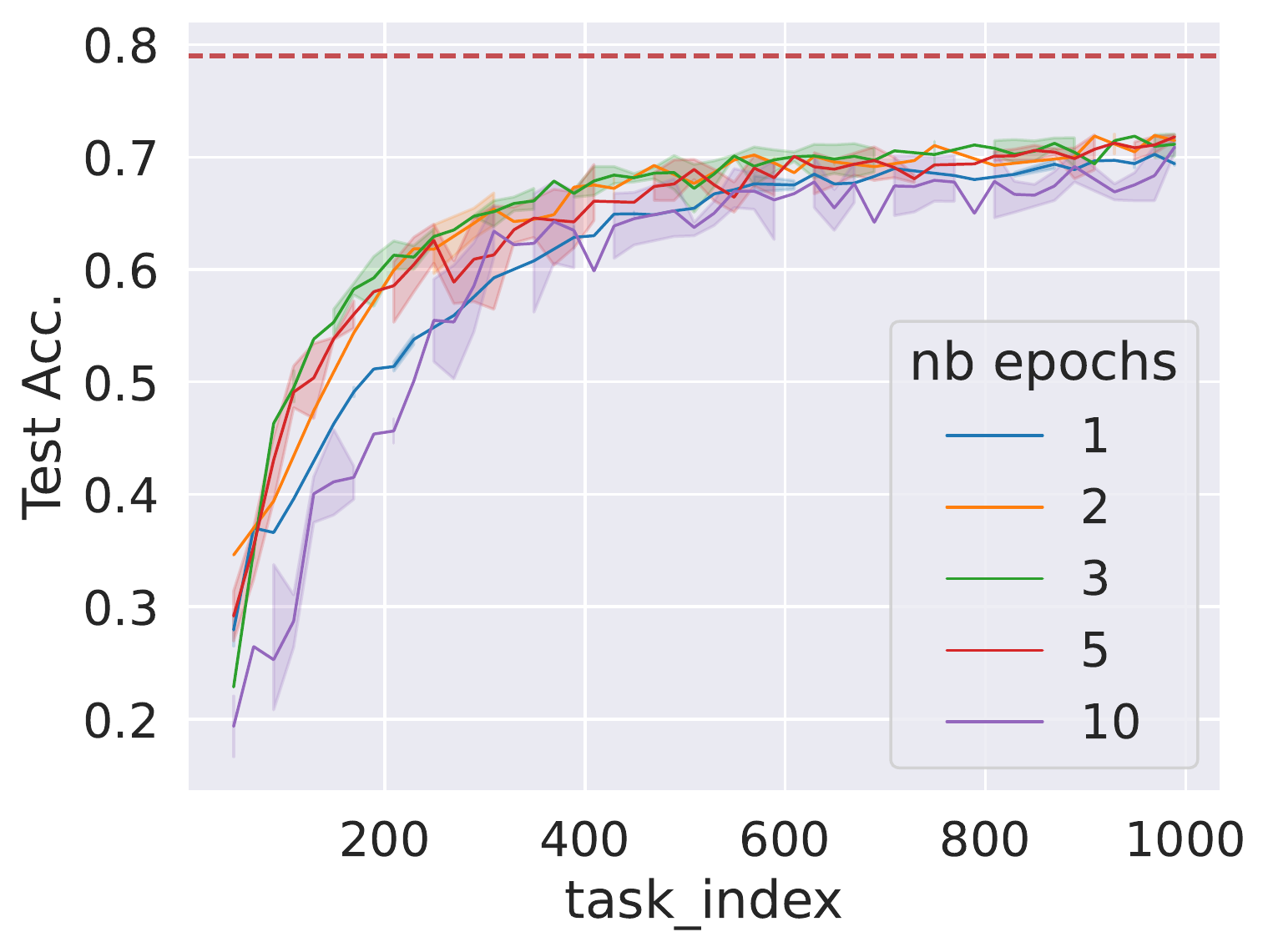}
    \end{subfigure}
    \begin{subfigure}[b]{0.49\linewidth}
    \centering
    \includegraphics[width=\linewidth]{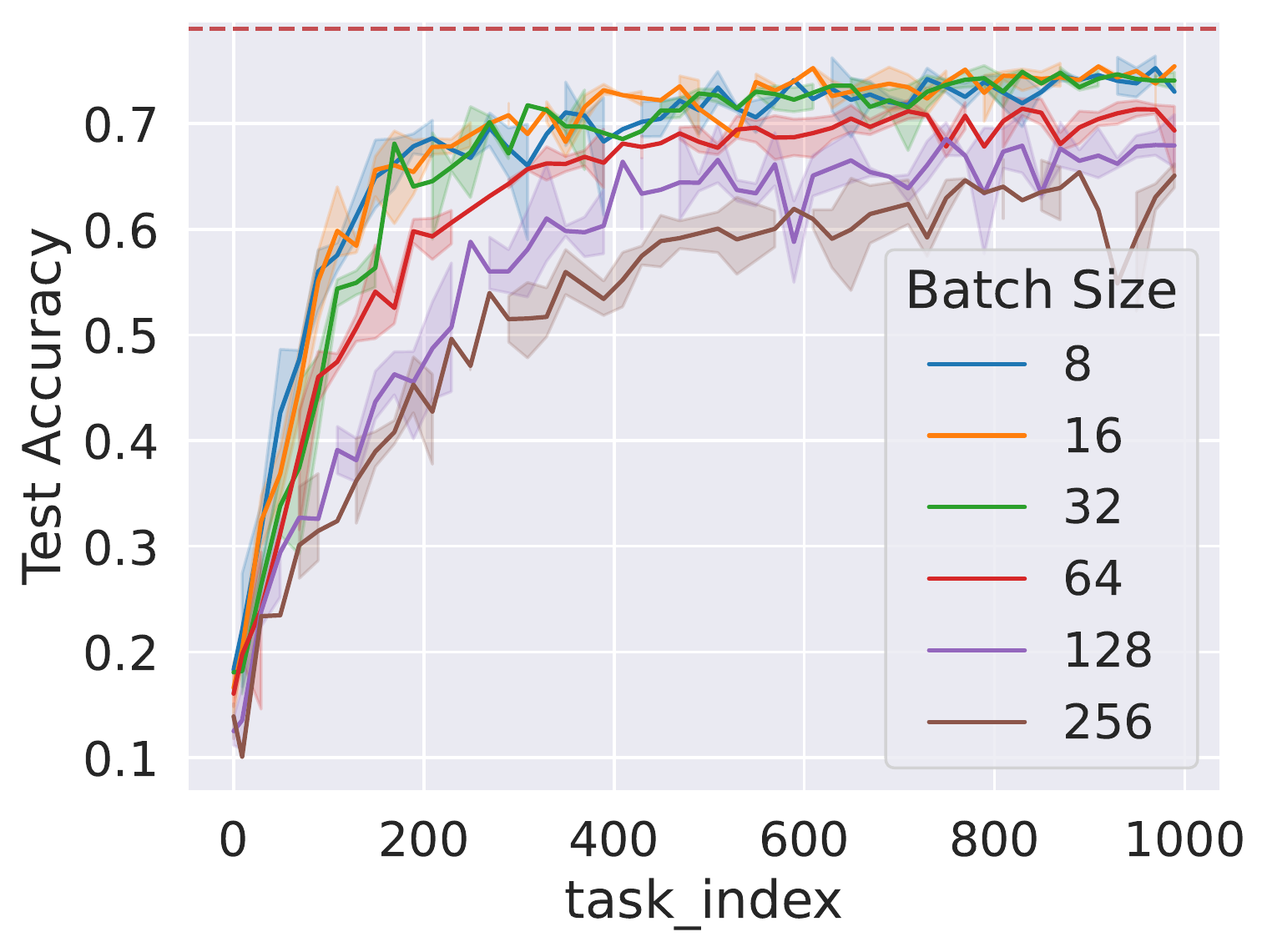}
    \end{subfigure}
    \caption{\modif{Test acc. on 1k tasks vs epochs per task and batch size. Performance may improve with more gradient steps per task but too many gradient steps on each task might lead to performance loss. \protect\tikz[baseline]{\protect\draw[red,line width=0.2mm, dashed] (0,.5ex)--++(0.5,0) ;} IID baseline.}}
    \label{ap:fig:learning-forgetting}
\end{figure}

\subsection{KL divergence \Scole vs IID.}
\label{ap:sec:kl}

\textbf{Measuring divergence to IID.}

In IID training, samples are drawn identically and independently from a static distribution. 
Hence, to break the IID assumption, we use the \Scole{} framework presented in \cref{sec:framework}, which constrains the sampling to be on a subset of classes that change through time. Hence, samples will not be identically and independently sampled.
To estimate how the \Scole{} framework differs from IID training, we compute the Kullback Leibler divergence between the IID and \Scole{} distribution and find that for $N$ classes and $C$ classes per task, $D_{KL}\big(p(X,Y|S_t)||P_{iid}(X,Y)\big) = \log \frac{N}{C}$. 

For a given task $t$, the KL-divergence between the data distributions is:
\begin{equation}
\begin{split}
    &D_{KL}\big(p(X,Y|S_t)||p_{iid}(X,Y)\big)\\
    & = \sum_{x\in X,y\in Y} p_t(x,y|S_t) \log\frac{p_t(x,y|S_t)}{p(x,y)} \\ 
    & = \sum_{x\in X,y\in Y} p(x|y)p(y|S_t) \log\frac{p(x|y)p(y|S_t)}{p(x|y)p(y)} \\
    & = \sum_{y\in Y}  p(y|S_t)\log\frac{p(y|S_t)}{p(y)} \sum_{x \in X} p(x|y) \\
    & = \sum_{y\in Y} \frac{\mathbbm{1}_{\{y\in S_t\}}}{C} \log \frac{N}{C} =
    \log \frac{N}{C}
\end{split}
\end{equation}           
Since $P_{\Scole{}}(Y))$ is uniform over elements in set $S_t$, $P_{\Scole{}}(Y=y)=\frac{1}{N}$. Since each of the $T$ tasks is equally probable, the expected divergence is as follows:
\begin{equation}
    \mathbb{E}[\log \frac{N}{C}] = \sum_{t=1}^{T} \frac{1}{T} \log \frac{N}{C} = \log \frac{N}{C}.
\end{equation}

This result shows that \Scole{} data distribution is different from an IID distribution and that we can control the KL divergence by changing $N$ and $C$.
Nevertheless, in the later experiments, instead of referring to the divergence, we will analyse the expected class frequencies $\nu^*_{class}$. Indeed, the frequency of classes is easier to estimate in

\subsection{Compute Estimation.}
\label{ap:sec:compute}

To estimate the compute overload of frequency replay ($21.28\%$) and full replay ($873\%$), we count the number of times any class is replayed. 
In a sequence of 2500 tasks with 10 classes per task, the finetuning approaches process the data of the equivalent of $2500\times10=25000$ classes. 
For frequency replay, the replay is triggered to replay a class $5320$ times within the 2500 tasks, while with full replay, it would be $193445$ times. 
Hence the estimated total compute of replay is $(25000+5320) \div 25000 = 1.2128$ time the compute of finetuning, and dense replay $(25000+193445) \div 25000 = 8.23$ times the compute of finetuning.

Samplewise, in total, fine-tuning approaches roughly processed $1.25\mathrm{e}{7}$ samples (2500 tasks with 10 classes per task, and approximately 500 samples per task), finetuning with frequency replay processed $1.5\mathrm{e}{7}$ while finetuning with full replay would have processed $1.09\mathrm{e}{8}$ samples.

To create random selection with the same compute budget, replay is realized by randomly and uniformly selecting a subset of classes in the buffer and replaying the associated data. The number of classes is selected, such that the overall compute growth by $21.28\%$ over simple finetuning.

\section{Additional Experiments}

\subsection{Reducing Variance of Expected Time between Repetition}
\label{ap:sub:prob_reduction}

\modif{In the original \Scole{} setup (~\cref{sec:framework}), the class distribution is stationary, hence sampling a new task is independent of what happened previously. This phenomenon can lead to a high variance in the number of intermediate tasks between two reoccurrences of a class. In order to reduce this variance, here we introduce a penalty factor $\gamma$ to penalize the resampling of classes that were sampled recently. Intuitively, we want to make the sampling probability of a particular class proportional to the time passed since it's last occurrence.}

\modif{We implement this idea in the following way: if a class is sampled in the current task, the probability of sampling this class in the next task is divided by $\gamma$. However, by applying this reduction the sum of the vector of probability becomes lower than 1 and a renormalization is applied to fix this.
The reduction + renormalization strategy decreases the probability of sampled classes and increase the probability for other classes.
One example, if we start with uniform distribution with $5$ classes we have a vector of probabilities $p=[0.2,0.2,0.2,0.2,0.2]$, if we sample two classes $[0,2]$ with a gamma factor of $2$, the unnormalized probability distribution becomes $p=[0.1,0.2,0.1,0.2,0.2]$ and after normalization $p=[0.125, 0.25 , 0.125, 0.25 , 0.25 ]$.
When applying this scheme sequentially we maximize the chance of resampling classes that did not appear recently, without changing the average expected time between two occurrences, therefore reducing the variance.}

\modif{Note that the higher $\gamma$ is, the closer we get to a fixed sequence of tasks the sequentially reoccurs, as in \cref{fig:fig_1}.}

\begin{figure}[!ht]
    \centering
    \includegraphics[width=0.5\linewidth]{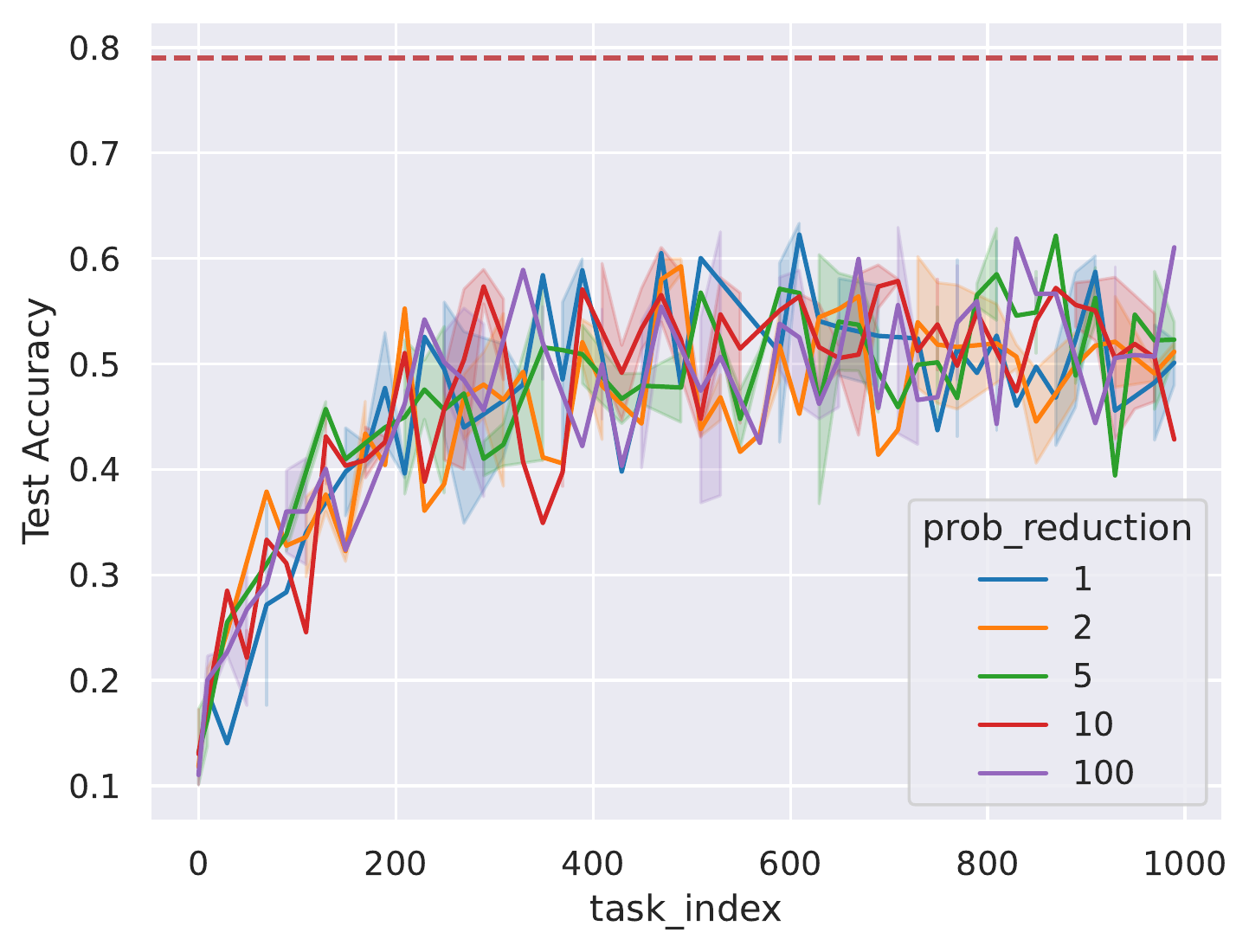}
    \caption{\modif{In this experiment, we investigated the effect of penalizing the reoccurrence of recent classes. A value of $\gamma=1$ indicates no penalization. Surprisingly, we found that this penalization had no discernible impact on knowledge accumulation.}.}
    \label{ap:fig:prob_reduction}
\end{figure}

\modif{We run experiments with various values of $\gamma$, ($\gamma=1$ is the same as the original \Scole{} setup) on CIFAR10 dataset with 2 classes per task ($N=10$, $C=2$).
The results of these experiments are shown in \cref{ap:fig:prob_reduction}, however, the penalization introduced by $\gamma$ did not seem to influence performance at all.
We conclude from this experience that the variance of the expected time between reoccurrences has a low influence on data accumulation, at least when the ratio $\frac{C}{N}$ is quite high. }

\subsection{Increasing the number of classes per task}

\begin{figure}[!ht]
    \centering
    \includegraphics[width=0.5\linewidth]{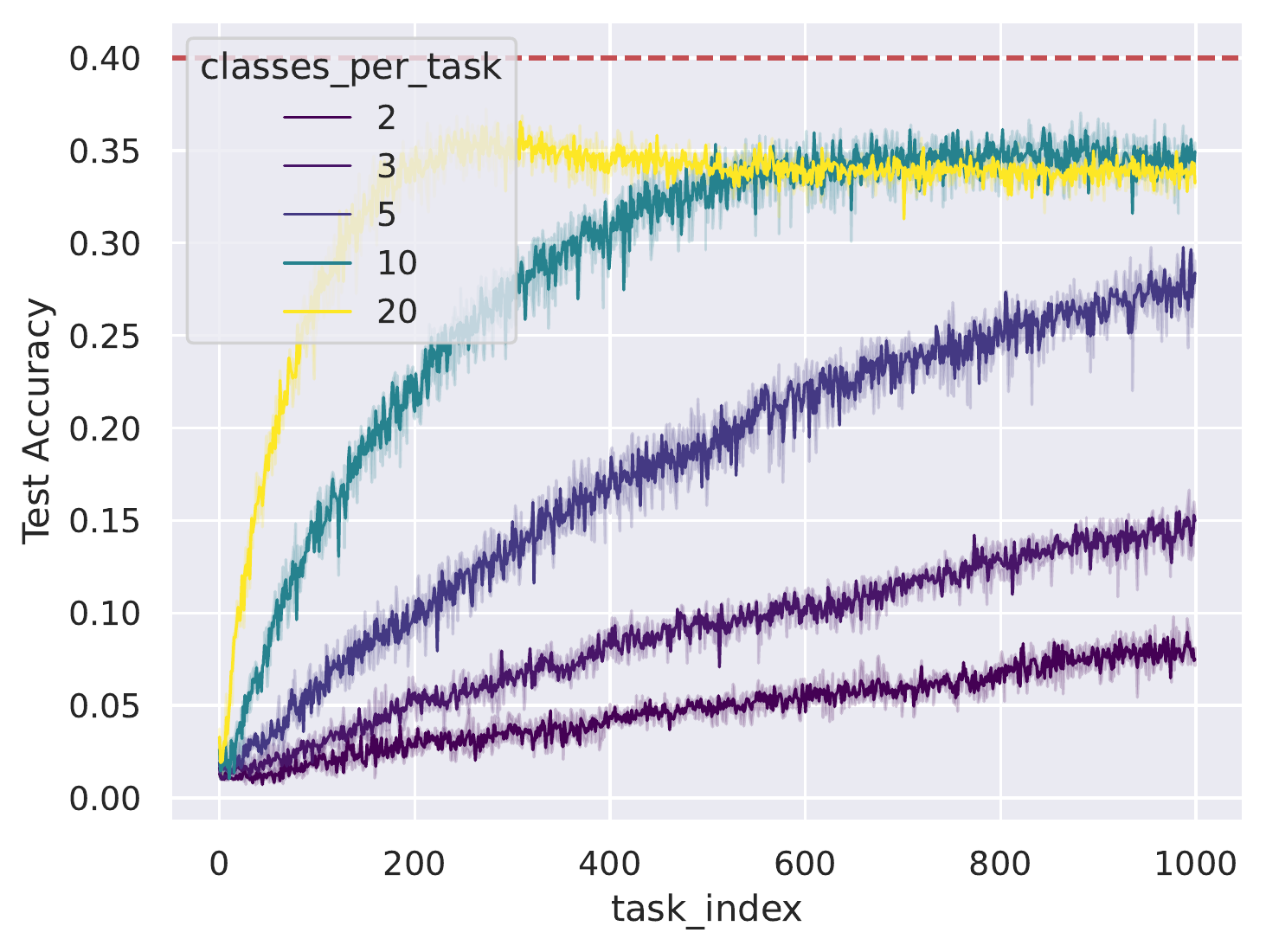}
    \caption{Growing the number of classes per task within a task in the full CIFAR100 dataset.}
    \label{ap:fig:growing_classes_per_tasks}
\end{figure}

\cref{ap:fig:growing_classes_per_tasks}, shows that our training framework also works when the number of classes per task increases. The more classes in the task, the faster the learning curve is. Here, we sample tasks from the entire CIFAR100 dataset.

\subsection{Making the model deeper}

\begin{figure}[!ht]
    \centering
    \includegraphics[width=0.5\linewidth]{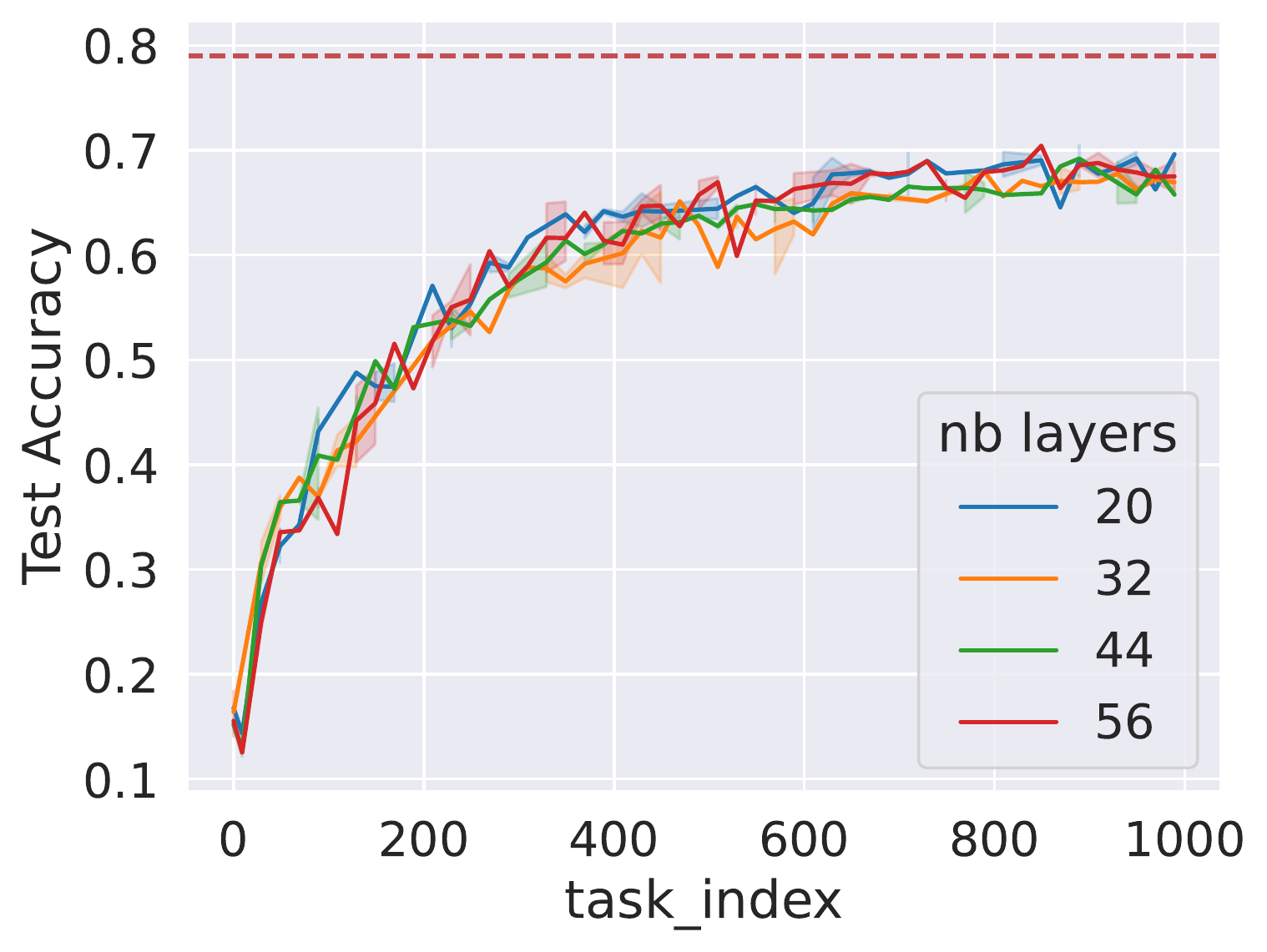}
    \caption{Growing the number of layers in the resnet model. \protect\tikz[baseline]{\protect\draw[red,line width=0.5mm, dashed] (0,.8ex)--++(1,0) ;} line represent IID performance with resnet22.}
    \label{ap:fig:going_deeper}
\end{figure}

In these experiments, we train on CIFAR10 with binary tasks and classes sampled uniformly. \cref{ap:fig:going_deeper} shows that depth has a low impact in our experiments.  

\subsection{Structuring the Task Sequence}
\label{ap:sub:structure}

\begin{figure}[!ht]
    \centering             
    \includegraphics[width=0.5\linewidth]{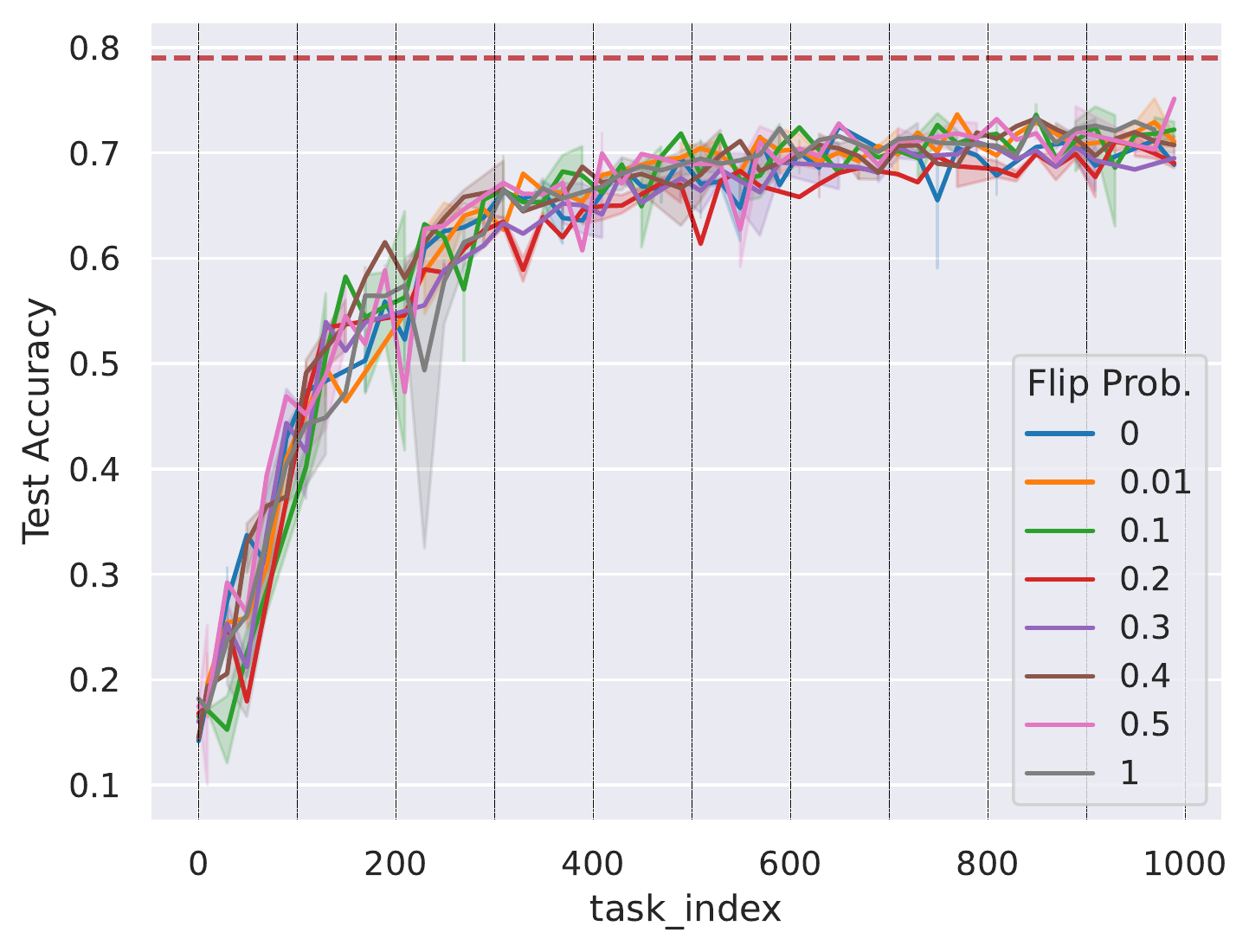}
    \caption{Comparison of several sequences of tasks with a default structure modified by a random flip of classes at each task. The scenario created with CIFAR10, 2 classes per task. The randomization of tasks is not critical for knowledge accumulation.}
    \label{ap:fig:structure}
\end{figure}

\textbf{Setting:} In this experiment, we want to evaluate the role of randomization of classes within tasks. In other words, we try to answer the following question: Is it important that classes are randomly sampled when building tasks?
For this, we start with a fixed sequence of binary classification tasks. This sequence is built so that all possible pairs of classes exist and occur only once. By default, training is achieved by repeating training on this fixed sequence of tasks until the end of the full sequence of tasks. We compare this baseline with the same sequence of tasks, but for each task, we set the probability $p$ that each class is flipped by random to another class.
The classes of the initial sequence of tasks is $[0,1] \xrightarrow[]{} [0,2]  \xrightarrow[]{} [...]  \xrightarrow[]{} [1,2] \xrightarrow[]{} [1,3] \xrightarrow[]{} [8,9] \xrightarrow[]{} [0,1] \xrightarrow[]{} [...]$.
The values of $p$ are $0, 0.01, 0.1, 0.2, 0.3, 0.4, 0.5, 1$. We can see that $p=1$ is the same as in the default scenario.

\textbf{Results:} The results presented in \cref{ap:fig:structure} show that having a fixed sequence of tasks instead of a randomized one does not reduce knowledge accumulation in our setting.

\subsection{Limiting the Possible Pairs of Classes in Tasks}
\label{ap:sub:pairs}

\begin{figure}[!ht]
    \centering             
    \includegraphics[width=0.5\linewidth]{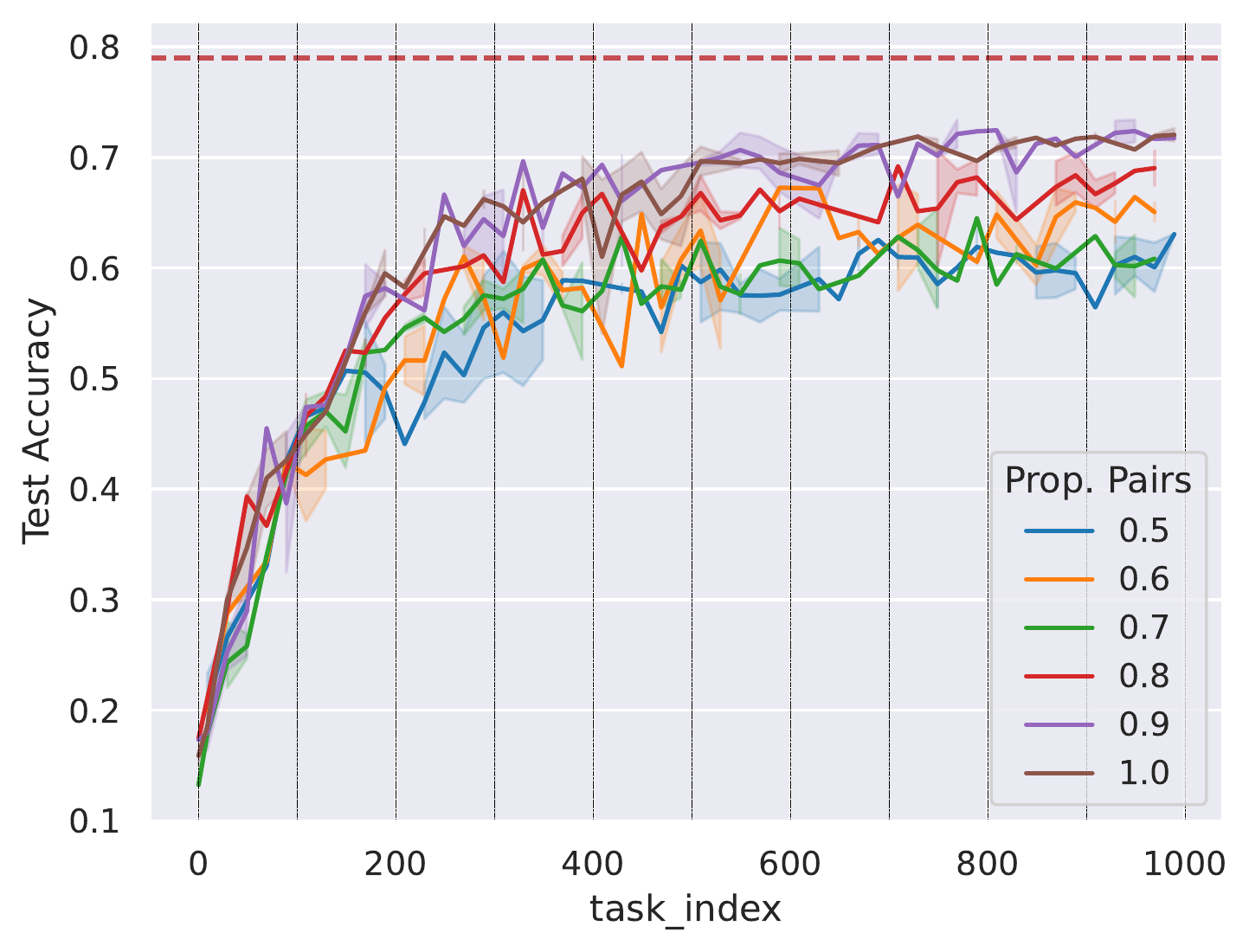}
    \caption{Comparison of \Scole{} scenario when selecting only a subset of all possible pairs of classes within tasks. We vary the proportion of pairs kept and plot test accuracy.}
    \label{ap:fig:couples}
\end{figure}

\textbf{Setting:} In this experiment, we want to evaluate how important it is that all possible pairs of classes exist within the full sequence of tasks. For this, we start from the list of all possible tasks and select only a subset of them. When building the task sequence, we only select pairs of classes from the selected list.

\textbf{Results:} The results presented in \cref{ap:fig:couples} show that the presence of all possible pairs of classes within the sequence of tasks plays an important role. In fact, without replay to learn discriminative features between two classes, classes must be on the same task \citep{lesort2019regularization}.

\section{Scenario Implementation}
\label{ap:sec:implementation}
\begin{figure}[!ht]
    
        \begin{python}
        # num_classes: the total number of classes
        # classes_per_tasks: number of tasks per class (2 by default)
        # probability: vector defining probability of sampling each class for a task (Uniform by default)
        # nb_epochs: epochs of training per task (1 by default)
        
        import numpy as np
        from continuum.scenarios import ClassIncremental
        from continuum.datasets import CIFAR10
        
        scenario = ClassIncremental(CIFAR10(config.data_dir, train=True), nb_tasks=nb_classes)
        test_set = CIFAR10(config.data_dir, train=False).to_taskset()
        
        for task_index in range(num_tasks):
            classes = np.random.choice(np.arange(num_classes), p=probability, size=classes_per_tasks, replace=False)
            
            # create taskset with only selected classes
            taskset = scenario[classes]
            for epoch in range(nb_epochs):
                # train the model on "taskset" data
                [...]
            # test the model on the full test set
            [...]
        \end{python}
        
\caption{Pseudo-Code using continuum \citep{douillard2021continuum} to control the distribution imbalance in classes.}
\label{code:implementation}
\end{figure}

The implementation presented in \cref{code:implementation} proposes a static version of the scenario. However, the \say{probability} distribution can be modified through the task sequence to create never-ending drifts or cyclic drifts in the class distribution or simply to change the balance of the class distribution.

\section{Mixture of class frequency}
\label{ap:sec:entropy_decrease}

\begin{figure}[!ht]
    \centering
        \begin{python}
        # num_classes: the total number of classes
        # lambda: hyper-parameter = 1/num_classes
        # entropy_decrease: parameter that control the scale of the imbalance
        
        import numpy as np
        prob_vec = np.ones(num_classes) / num_classes
        prob_vec = prob_vec - (1/num_classes) * np.arange(num_classes) * lambda
        prob_vec /= prob_vec.sum()
        prob_vec = prob_vec**entropy_decrease / (prob_vec**entropy_decrease).sum()
        
        # we shuffle the vector so for each experiment the imbalance is not the same
        np.random.seed(config.seed)
        np.random.shuffle(prob_vec)
        
        for task_indef in range(num_tasks):
            selected_classes = np.random.choice(np.arange(num_classes), p=prob_vec, size=classes_per_task, replace=False)
            # the we can create the task and train the model
            [...]
        \end{python}
        
\caption{Pseudo-Code to create imbalance frequency of appearance.}
\label{code:entropy_decrease}
\end{figure}

To change the entropy of the class distribution, we start with a uniform vector of probabilities $u$.
For each class, $u$ gives the probability of this class being sampled for a task. 
To create an imbalance in class probabilities, we slightly modify this vector using $u' = u - \frac{1}{C^2} * numpy.arange(C)$, with $C$ the number of classes. The modification is quadratic with the number of tasks that cover a wide range of occurrence frequency.
To increase the imbalance in the distribution, we multiply $u'$ by itself $d$ times. The complete experimentation tests the values of $0,1,2,5$ and $10$. Note that $d=0$ means that the distribution is uniform. \cref{ap:sec:entropy_decrease} shows the Python implementation and the probability vectors for each $d$).

\begin{figure}[!ht]
    \centering             
    \includegraphics[width=0.5\linewidth]{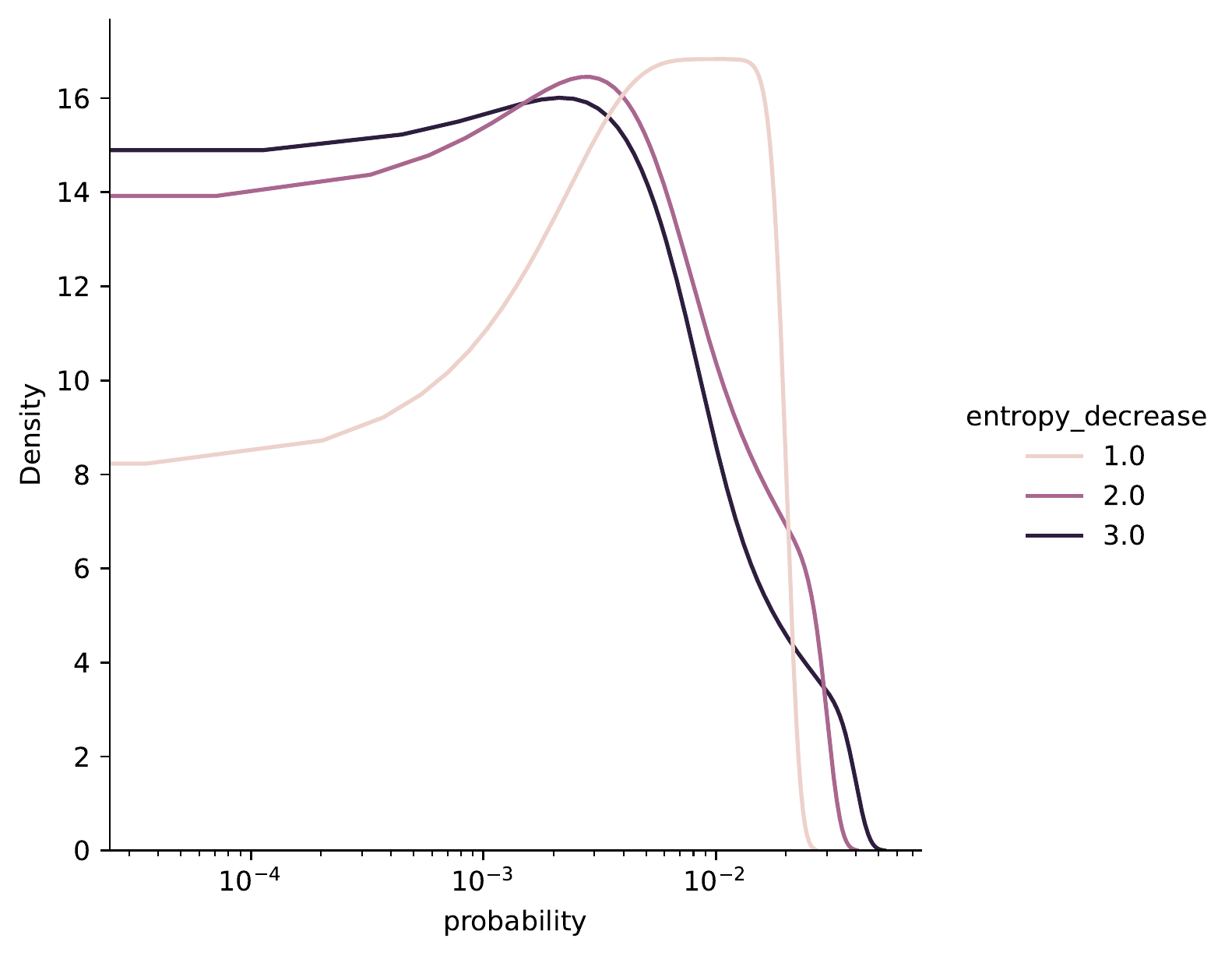}
    \caption{Density distribution of log probabilities of sampling for various entropy decrease factors.}
    \label{ap:fig:density_probabilities}
\end{figure}

\begin{figure}[!ht]
    \centering             
    \includegraphics[width=0.5\linewidth]{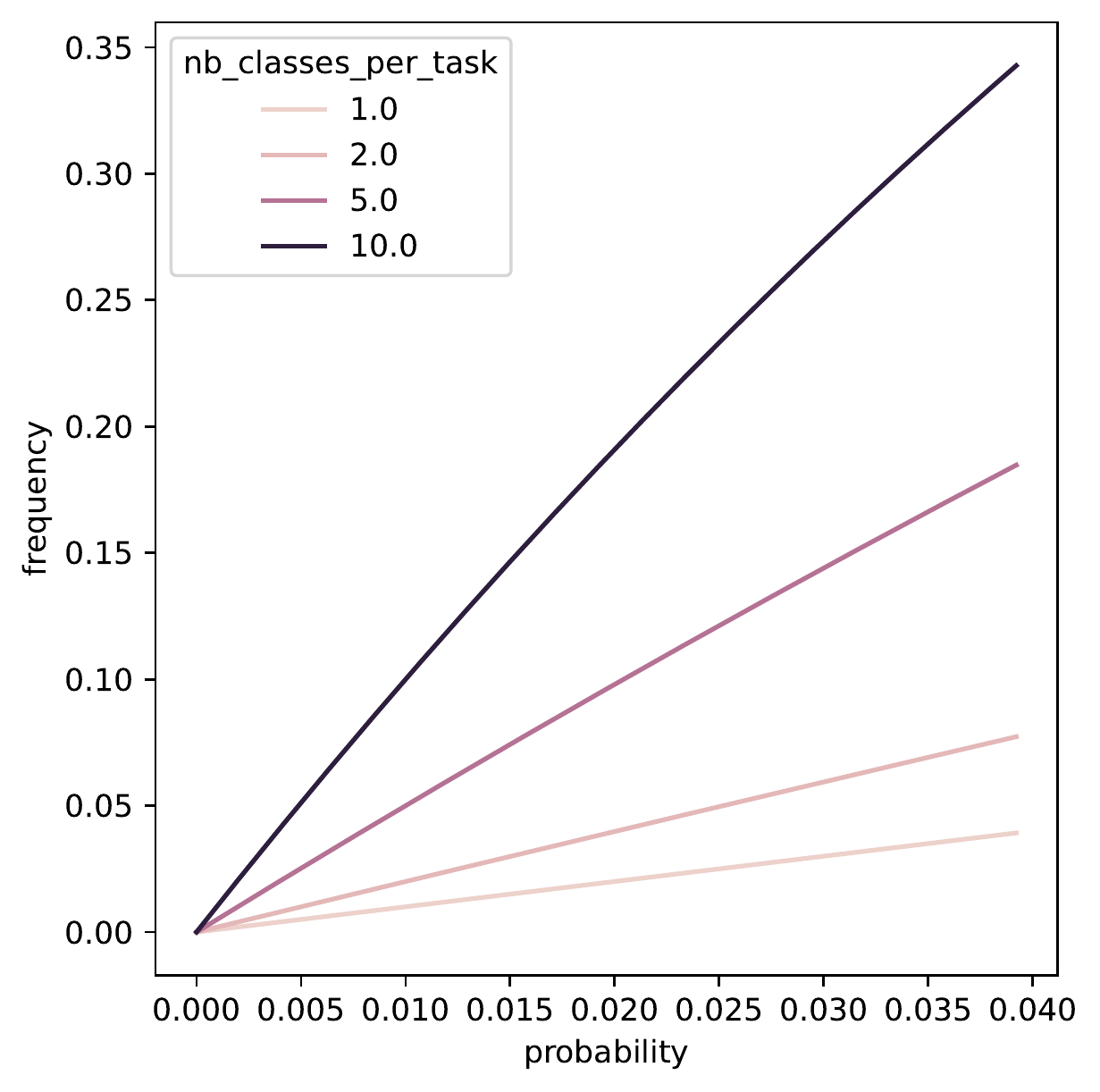}
    \caption{Probility of occurrence vs frequency of occurrence for a various number of classes per task. The number of classes per task grows the frequency of occurrence.}
    \label{ap:fig:proba_vs_freq}
\end{figure}


\section{Compute}
\label{ap:sec:compute}

This project was realized with the use of internal clusters. Each run was performed with a single GPU, mostly NVIDIA GeForce RTX 2070, Quadro RTX 8000, and Tesla V100-SXM2-32GB. The total amount of time required for the runs is 202 days.

\section{Architectures Neural Network First Experiment}
\label{ap:sec:cnn}

The architecture of the convolutional neural network used in \cref{sec:preliminary}.

\begin{figure}[!ht]
    
        \begin{python}
        import torch.nn as nn
        
        relu = nn.ReLU()
        conv1 = nn.Conv2d(1, 10, kernel_size=5)
        conv2 = nn.Conv2d(10, 20, kernel_size=5)
        maxpool2 = nn.MaxPool2d(kernel_size=2)
        fc1 = nn.Linear(320, 50)
        head = nn.Linear(50, 10)
        
        # Forward pass with pytorch
        # x dimension is [1,28,28]
        x = relu(maxpool2(conv1(x)))
        x = relu(maxpool2(conv2(x)))
        x = relu(fc1(x))
        x = head(x)
        
        \end{python}
        
\caption{Pseudocode describing the architecture used for experiments in \cref{sec:preliminary}}
\label{code:architecture}
\end{figure}

\section{Forgetting}
\label{ap:sec:fg_details}

In long task sequences, it becomes difficult and costly to track the forgetting on all tasks seen so far, hence we estimate forgetting by looking locally at how learning a new task makes the model forget the one just before.
We calculate local forgetting, which is the amount of forgetting in a task-induced by learning the next task.
Note: we only compute forgetting with non-overlapping classes between two tasks.

\begin{equation}
    F_{local}(t) = \frac{1}{N-C} \sum_{j \notin \mathcal{Y}_t} A_{t, y=j} - A_{t-1, y=j}
    \label{ap:eq:local_F}
\end{equation}
With $\mathcal{Y}_t$ the set of classes in task $t$ and $A_{t, y=j}$ the accuracy realized on class $j$ at task $t$.
$F_{local}(t)$ averages the forgetting generated in classes that are not in the current task.
Then, total forgetting $F$ averages local forgetting on tasks seen to far and is computed as:

\begin{equation}
    F = \frac{1}{T-1} \sum_{i=1}^{T} F_{local}(t)
    \label{ap:eq:F}
\end{equation}

\section{Effect of long-term distribution shifts}
\label{ap:forgetting}

In the experiments presented so far, we sampled classes for each task from the same distribution $p(S_t;C)$. 
Here we create \Scole{} scenarios with a shift over time in the class distribution and assess the accumulation of knowledge under long-term shifts --- shifts that persist for over several hundreds of tasks. Those shifts may result in very large or infinite $\tau_{class}$ for some classes.
We evaluate knowledge retention and interference by designing three different shift patterns and 
evaluate models with increasing width to access whether increasing the width can slow down forgetting \citep{mirzadeh2021wide}. 

\subsection{Knowledge retention with long-term class shift}

We assess the capabilities of deep neural networks to maintain correct prediction on classes that stop appearing. This evaluation of knowledge retention is strict in comparison to related works~\citep{fini2022self, davari2022probing} that evaluate knowledge retention by linear probing latent representation. In our setup, knowledge retention in lower layers is not sufficient, and the model also has to maintain knowledge from feature extraction in lower layers to the decision boundaries in the last layer.

\textbf{Setting:} We train on a scenario with $C=2$ and uniform $p(Y|S_t)$  
on CIFAR10. 
We start the training with all classes, and after 500 tasks, we remove half the classes from the class distribution.
In contrast to standard CL, where distribution shift is usually caused by adding classes, here we remove classes. In such a scenario, the forgetting behaviour should be smooth because the error on remaining classes should be low. 
%
\textbf{Results:} \cref{fig:forgetting} (left) shows that even if no new data is introduced to the learner, that is, no interference with old knowledge is possible, the model can still forget when a subset of already learned classes is no longer trained on. Interestingly, in this setup forgetting is slow and not catastrophic, and knowledge persists during several hundreds of tasks. 
Moreover, we clearly observe that growing the width of the model increases knowledge retention to the point that it looks like the model perfectly memorized removed classes for the maximum model size. 

\subsection{Knowledge retention with class substitution} 
\label{sub:incremental}
Similarly to the previous section, we investigate a setting with an abrupt shift in the class distribution,  however instead of removing classes (shrinking the domain of $p(S_t;C)$), we replace existing classes with new (shift the domain of $p(S_t;C)$). This allows us to investigate the interference and forgetting dynamics that both cause performance drop in this setting. 
The goal is to assess whether observed knowledge retention 
can help to slow down forgetting. 

\textbf{Setting:}
First, 500 tasks are generated from the first 5 classes of CIFAR10 (first period), and the second 500 tasks are generated from the remaining 5 classes (second period). There is then no overlap between the classes in the first and second periods. Also here we test models with various widths.
\textbf{Results:}
The results in \cref{fig:forgetting} (right) show that this sudden class shift creates disturbances in the training process as in classical CL scenarios. Moreover, in the second period of training, the model struggles more to learn the tasks than during the first period, meaning the first period does not provide good initialization for later tasks and that the forward transfer is limited in such a regime. This result corroborates the results of \citet{ash2020warm}, who witnessed a similar phenomenon in a transfer setting.
Similarly to the previous section, observe that wider models can better resist catastrophic forgetting. Still, this difference appears less clear, and it could be due to the better knowledge retention of first classes.

\subsection{Cyclic Shifts}
\label{ap:sub:cyclic}

\begin{figure}
    \centering
    \begin{subfigure}[b]{0.4\linewidth}
    \includegraphics[width=\linewidth]{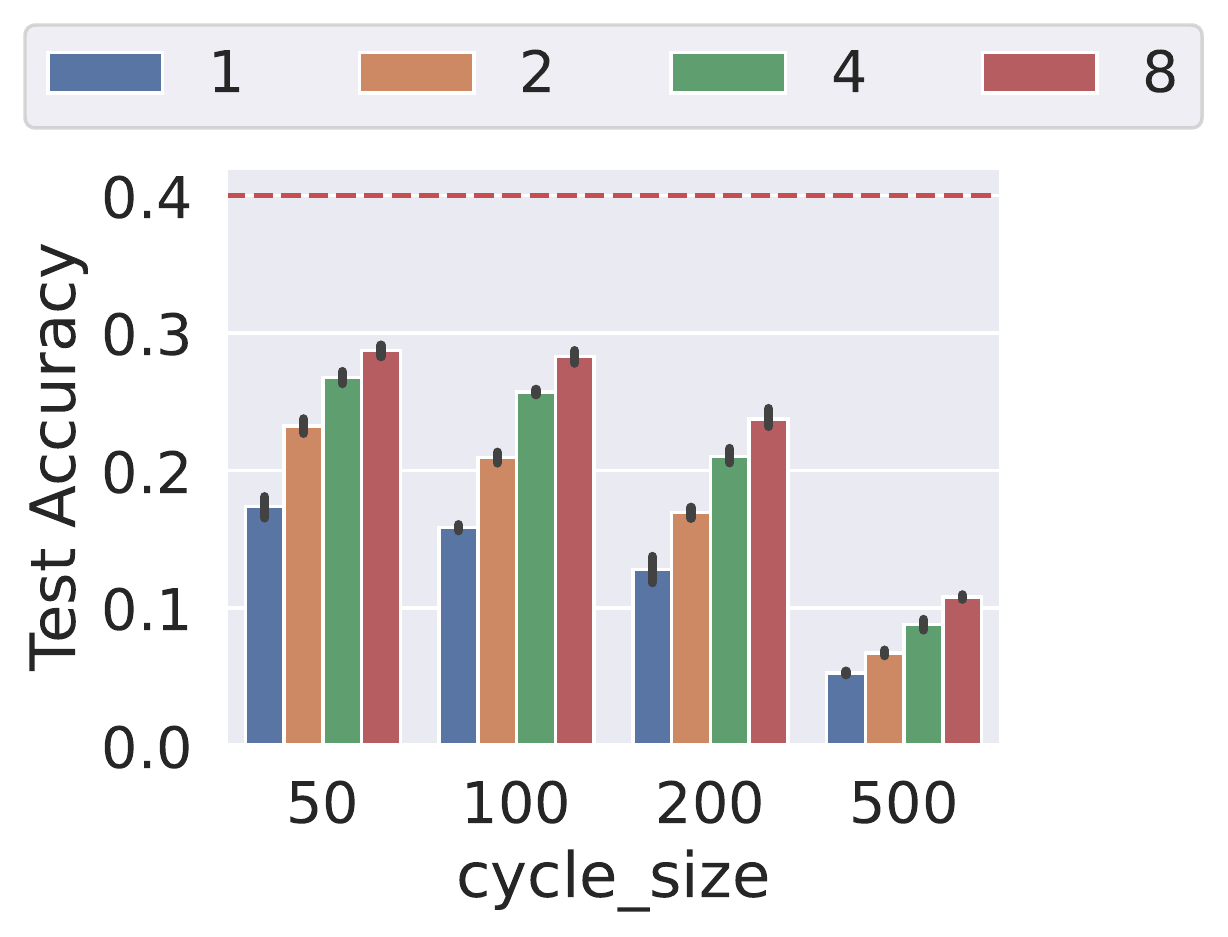}
    \end{subfigure}
    \begin{subfigure}[b]{0.4\linewidth}
    \includegraphics[width=\linewidth]{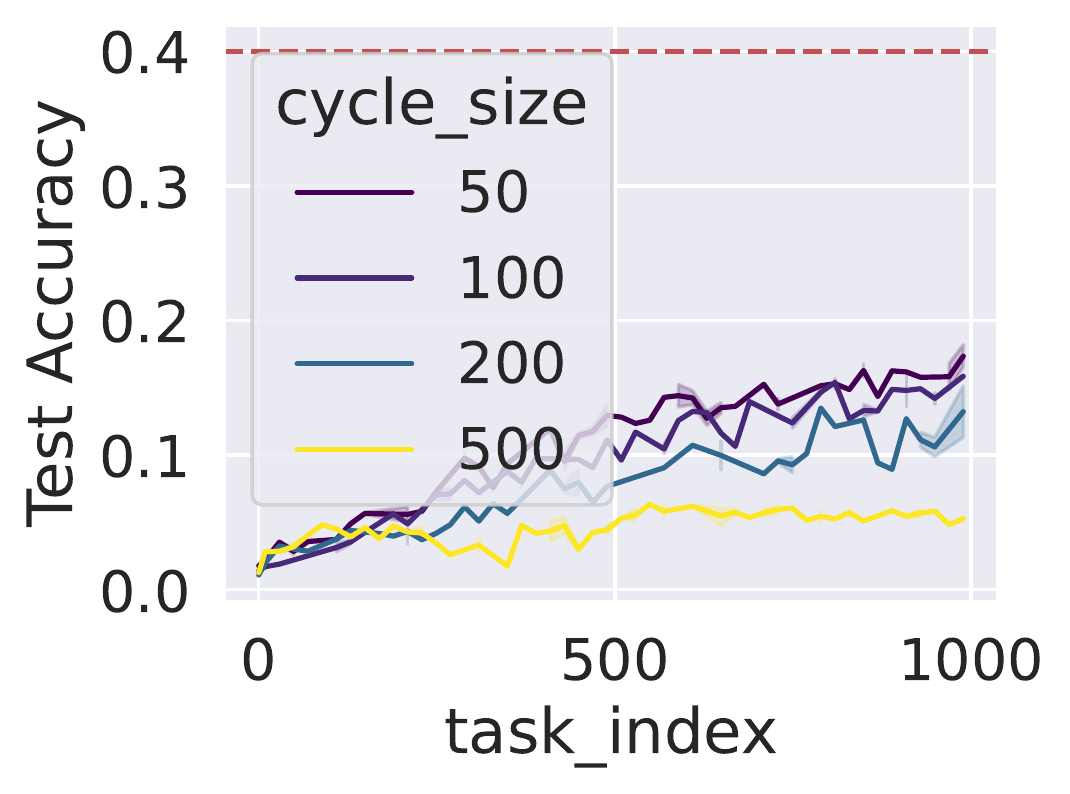}
    \end{subfigure}
    \caption{Cyclic Shifts Experiments: (left) shows the test accuracy averaged over the 10 last tasks for all cycle and model sizes. (right) the effect of cyclic shifts for the smaller model.}
    \label{fig:cyclic_exps}
\end{figure}

Here we aim to assess whether knowledge retention in DNNs observed previously persists when we fix $\tau_{class}$ to be equal for all classes but varying number of expected SGD updates $\overline{u}_{class}$ before repetitions. To this end, we let the distribution $p(S_t;C)$ follow a cyclic pattern.  

\textbf{Setting:}   
We define a shifting window of $W$ classes as the domain of $p(S_t;C)$ for $\frac{N}{\lambda}$ tasks after which it is shifted by one class or more depending on $N$ and $\lambda$. Here $\lambda$ is the cycle size, i.e. $W$ is exactly the same for every $\lambda$ tasks. 
For example, if the class subset $W_t$ before is $[0,1,2]$, after a shift, it will become $[1,2,3]$. After $\lambda$ tasks, it will be again $[0,1,2]$. 
We use the full CIFAR100 dataset ($N=100$) with 5 classes per task, the window size is $W=10$, and the cycle size is $\lambda \in [50,100,200,500]$, higher $\lambda$ leads to higher $\overline{u}_{class}$. We choose CIFAR100 because it has more classes than CIFAR10 and allows the creation of shifts over a longer period.
In this experiment, when a class leaves the subset window, it needs $\lambda-W$ tasks to return leading to equal $\tau_{class}$ for all classes. 
  
\textbf{Results:}
We can see in \cref{fig:cyclic_exps} that an increased shift period makes training harder, i.e., better learning happens if classes reoccur more frequently. Even with the largest period, the model still progresses systematically over time, as seen in \cref{fig:cyclic_exps} (bottom). 
Consistently with other experiments, wider models result in better performance, as shown in \cref{fig:cyclic_exps} (top). 
These experiments show that SGD-trained DNNs are also capable of long-term retention and can still accumulate even if they do not see some classes for a long period of time. 

{\bf Conclusion. } Forgetting still happening in long sequences of tasks with long-term distribution shifts is expected. However, experiments in this section show that models are capable of long-term knowledge retention, enabling knowledge accumulation even when data is not seen for a long time.
Our results are in line with the findings of \citet{mirzadeh2021wide}: Wider models forget less in incremental scenarios. Further, we have also shown that the widest models are capable of almost perfect knowledge retention and that with long-term distribution shifts such as cyclic shifts, deep neural networks can accumulate knowledge. They can memorize and reuse knowledge from classes not seen since more than 200 hundred tasks.

\section{Experiments with Data augmentation}

\subsection{Samples}
\label{ap:DA-samples}

\begin{figure}
    \begin{subfigure}[b]{0.19\linewidth}
    \centering
    \includegraphics[width=\linewidth]{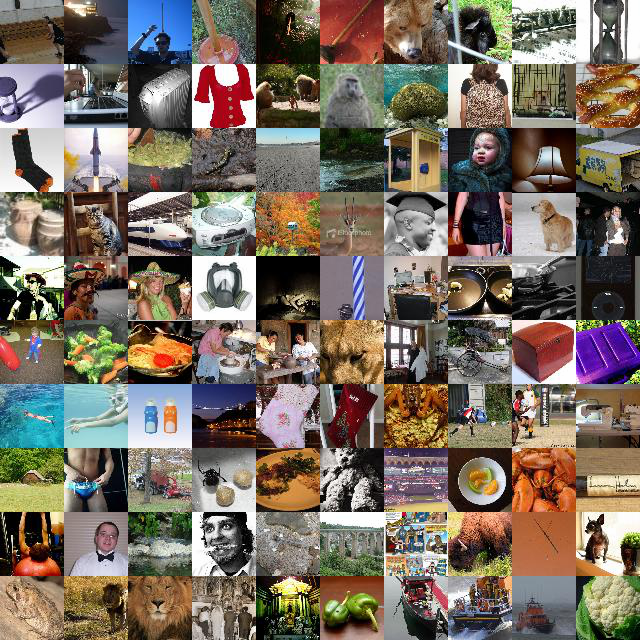}
    \caption{0 - Original}
    \label{fig:None}
    \end{subfigure}
    \begin{subfigure}[b]{0.19\linewidth}
    \centering
    \includegraphics[width=\linewidth]{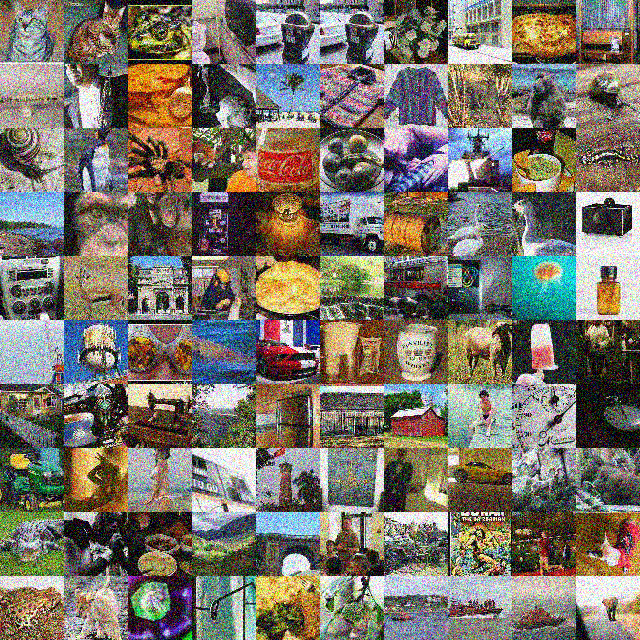}
    \caption{ 1 - gaussian\_noise}
    \label{fig:gaussian_noise}
    \end{subfigure}
    \begin{subfigure}[b]{0.19\linewidth}
    \centering
    \includegraphics[width=\linewidth]{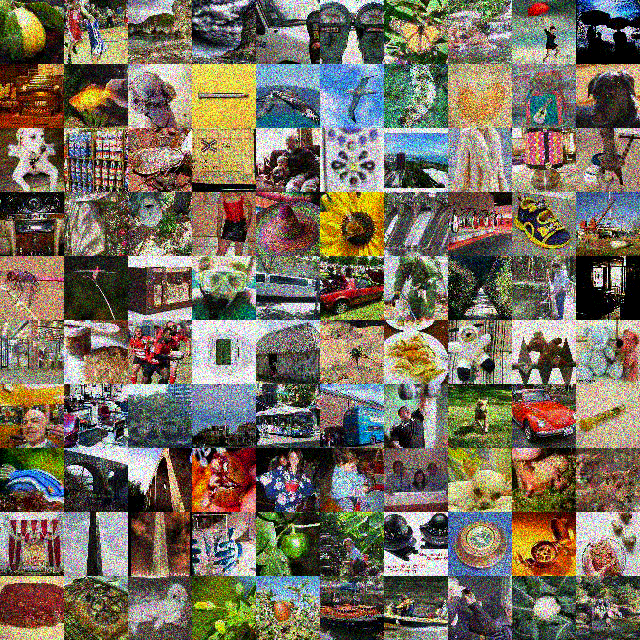}
    \caption{2 - shot\_noise}
    \label{fig:shot_noise}
    \end{subfigure}
    \begin{subfigure}[b]{0.19\linewidth}
    \centering
    \includegraphics[width=\linewidth]{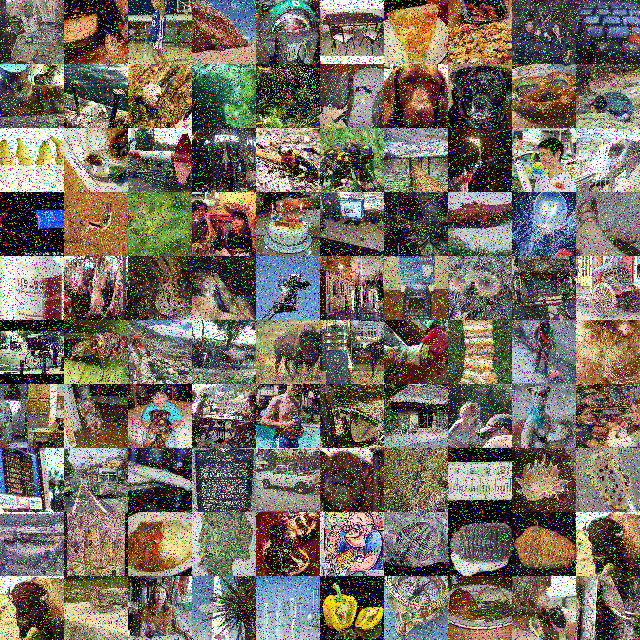}
    \caption{3 - impulse\_noisee}
    \label{fig:impulse_noise}
    \end{subfigure}
    \begin{subfigure}[b]{0.19\linewidth}
    \centering
    \includegraphics[width=\linewidth]{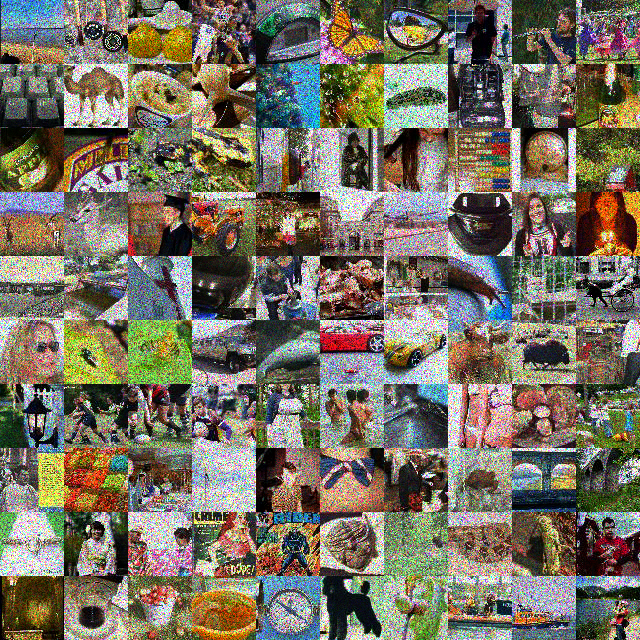}
    \caption{4 - speckle\_noise}
    \label{fig:speckle_noise}
    \end{subfigure}

    \caption{Example of data augmentation for experiments in \cref{sub:replay}. The data augmentation is here realized with severity 4 for better visual effect. However, experiments were conducted with severity 1 in order to not make images unrecognizable. }
\end{figure}

\subsection{Supplementary Experiments}
In most of our experimentation, when a class reappears, the exact same data is fed to the model. In this experiment, we want to investigate the influence of this. We compare training with exact same data and with data modified by random data augmentation.

\begin{figure}[!ht]
    \centering
    \includegraphics[width=0.45\linewidth]{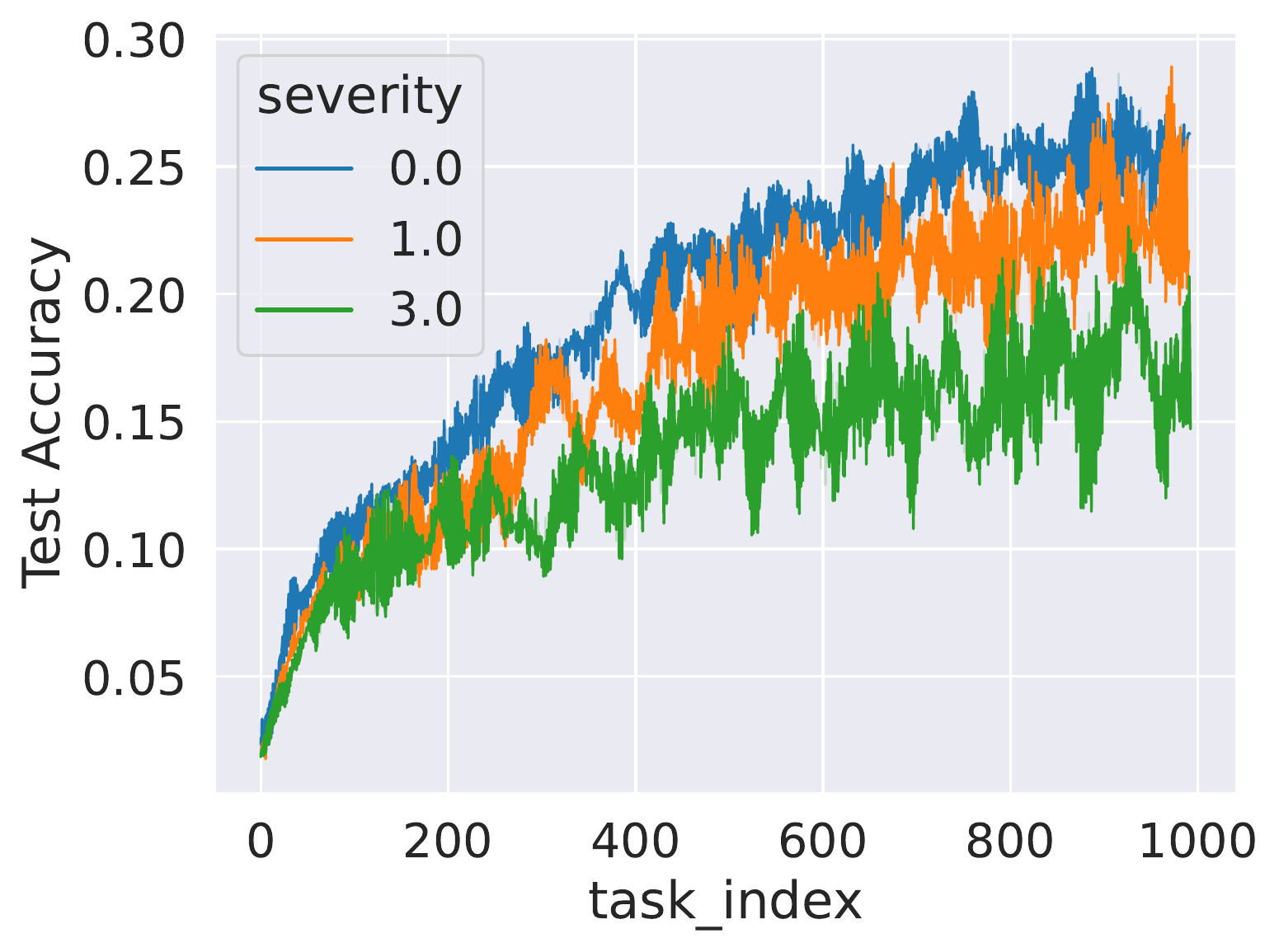}
    \caption{TinyImagenet 50/5, 1 epoch per task: Addition of random augmentation at each task with various severity. The augmentation slows down knowledge accumulation but does not prevent it.}
    \label{ap:fig:aumgentation}
\end{figure}

\begin{figure}[!ht]
    \centering
    \includegraphics[width=0.45\linewidth]{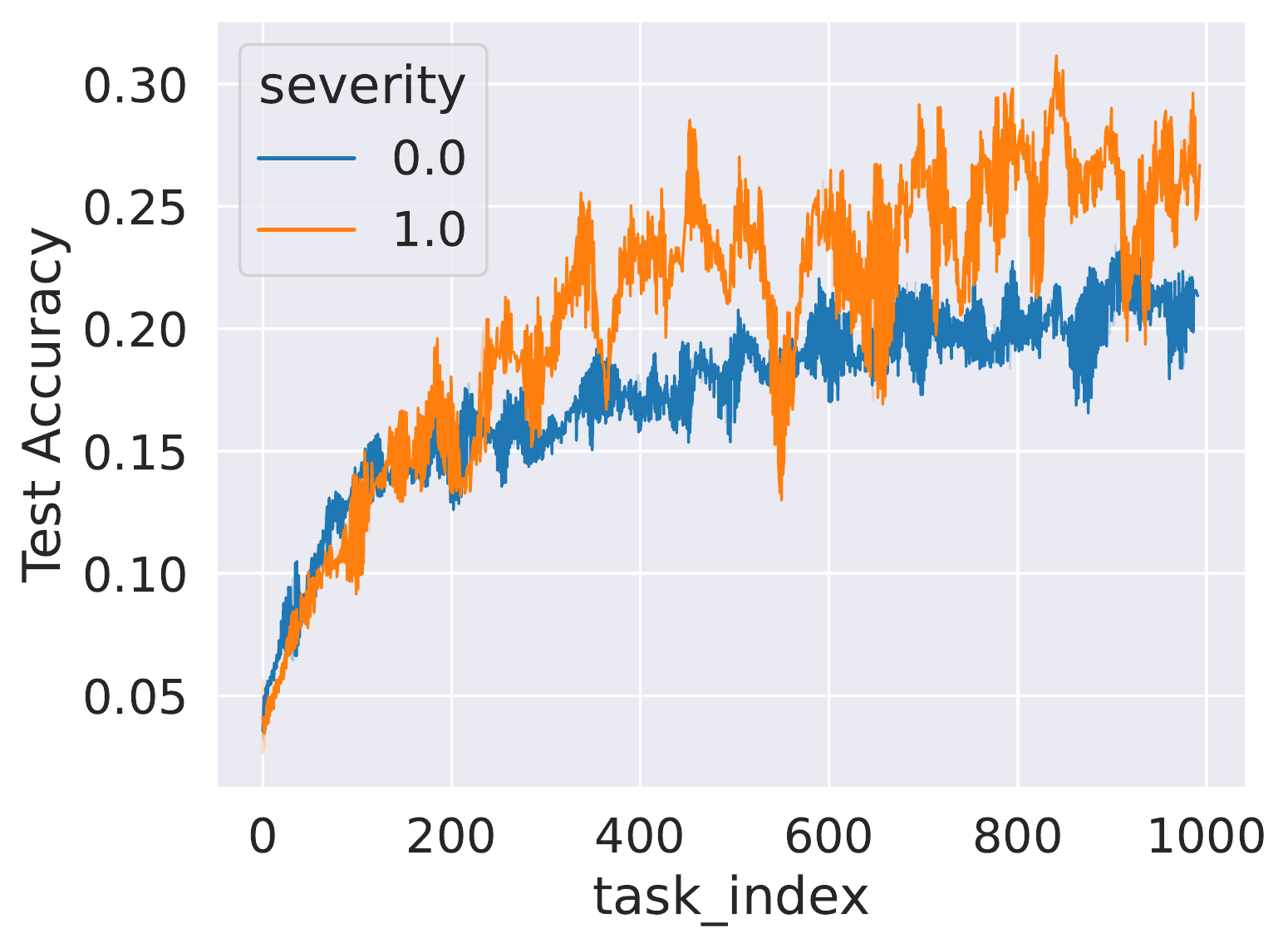}
    \caption{TinyImagenet 50/5, 5 epochs per task: when growing the number of epochs per task, a slight data augmentation improves results over exact data repetition.}
    \label{ap:fig:aumgentation_5}
\end{figure}

\textbf{Setting:} We use the augmentations proposed in \cite{hendrycks2019benchmarking} that simulate common perturbation and corruption proposed in images. At each task, a new augmentation is selected and applied with severity 1. We selected our augmentation among, \say{no corruption},
\say{gaussian noise}, \say{shot noise}, \say{impulse noise}, \say{speckle noise}, \say{gaussian blur}, \say{defocus blur}, \say{motion blur}, \say{zoom blur}, \say{fog}, \say{snow}, \say{spatter}, \say{contrast}, \say{brightness}, \say{saturate}, \say{elastic transform} and \say{glass blur}.
They minimized the chance of having the same exact data several times, and the augmentation applied to the data is variate and significant.
We use a subset of 50 of the TinyImagenet dataset with 5 classes per task.

\textbf{Results:} Our results are presented in \cref{ap:fig:aumgentation}. It shows that the augmentation applied to each image to minimize the chances of having the exact same images does not compromise the accumulation of knowledge. Moreover, \cref{ap:fig:aumgentation_5} shows that when growing the number of epochs per task, augmentation can improve performance over exact data repetition.

\end{document}